\documentclass[11pt]{article}

\usepackage[T1]{fontenc}
\usepackage{lmodern}
\usepackage[margin=1in]{geometry}
\usepackage{amsmath,amssymb}
\usepackage{array,booktabs,tabularx}
\usepackage{graphicx}
\usepackage{float}
\usepackage{xcolor}
\usepackage{microtype}
\usepackage[numbers,sort&compress]{natbib}
\usepackage{hyperref}
\usepackage{graphcore_report}

\graphicspath{{figures/}}
\newcolumntype{Y}{>{\raggedright\arraybackslash}X}

\newcommand{\ue}{UE5M3}
\newcommand{\uefp}{UE5M3 FP4}
\newcommand{\nv}{NVFP4}
\newcommand{\mx}{MXFP4}
\newcommand{\eformat}[1]{\texttt{#1}}

\title{UE5M3 FP4 Block Scaling for Stable Language Model Pretraining}
\reportsubtitle{A Simpler FP4 Pretraining Recipe with Delayed Tensor Scaling}
\author{Robert Hu, Carlo Luschi, and Paul Balanca\\[1.5mm]
  \mdseries\small Correspondence:
  \href{mailto:robert.stats.hu@gmail.com}{robert.stats.hu@gmail.com}}
\date{September 2026}
\hypersetup{
  colorlinks=true,
  linkcolor=GraphcoreInk,
  citecolor=GraphcoreCoralDark,
  urlcolor=GraphcoreCoralDark,
  pdftitle={UE5M3 FP4 Block Scaling for Stable Language Model Pretraining},
  pdfauthor={Robert Hu, Carlo Luschi, and Paul Balanca},
  pdfsubject={UE5M3 FP4 pretraining with delayed tensor scaling},
  pdfkeywords={FP4, UE5M3, NVFP4, microscaling, stochastic rounding, delayed scaling}
}

\begin{document}
\maketitle

\begin{abstract}
Stable 4-bit floating-point (FP4) pretraining is difficult because the E2M1
payload represents only a narrow range of magnitudes.  NVIDIA's Transformer
Engine \nv{} recipe addresses this with current-tensor scaling, a randomized
Hadamard transform (RHT), and bfloat16 (BF16) final layers, adding work outside
the FP4 matrix multiplications.  We instead pair E2M1 payloads with unsigned
E5M3 (\ue{}) block scales.  Their wider range permits periodic tensor scaling,
while our recipe applies selective stochastic rounding to backward gradients,
omits RHT, and uses FP4 in all eligible internal linears.

We pretrain a Nemotron-H 8B model for nearly 190 billion tokens.  Compared with
Transformer Engine \nv{}, the proposed block-16 recipe finishes with lower
final-window training loss and, under their respective quantized-inference
policies, lower validation loss measured as held-out negative log-likelihood.
Its quantized-inference downstream point estimates are also higher on all three
reported aggregates.  A native \nv{} execution ablation that jointly removes
RHT and the BF16 final-block exemption increases measured model-body token
throughput by 21.2\%.  These results demonstrate end-to-end software-emulated
\uefp{} pretraining with a simpler recipe and motivate native support for \ue{}
block scaling.
\end{abstract}

\section{Introduction}

On the GB200 platform evaluated here, supported FP4 Tensor Core operations have
four times the theoretical peak throughput of BF16~\citep{nvidia2025nvfp4}.
This makes FP4 attractive for both inference and pretraining.  Inference can quantize fixed weights and
calibrate scales after training; pretraining must repeatedly quantize changing
activations, weights, and gradients while preserving the optimization signal.
This report focuses on the latter problem: stable end-to-end \uefp{}
pretraining.

The use of FP4 can also reduce operand traffic because each payload element uses one
quarter as many bits as BF16, although block- and tensor-scale metadata reduce
the realized storage ratio.  The small E2M1 codebook makes pretraining
sensitive to scale selection, outliers, rounding, and accumulation.  Current
tensor-scale reductions, RHT operators, and BF16 final layers can stabilize
optimization, but each adds work outside the FP4 matrix multiplications.  We
ask whether the use of a wider block-scale format can retain stability with a simpler
recipe.

FP8 training is now well established~\citep{micikevicius2022fp8}.  FP4 is a
harder target.  The E2M1 FP4 payload used by \mx{} and \nv{} contains one sign
bit, two exponent bits, and one fraction bit; its largest magnitude is only 6.
Scaling is therefore required to represent the dynamic range of activations,
weights, and gradients.  Microscaling uses one scale per small
block---typically 16 or 32 elements---rather than one scale for the entire tensor~\citep{rouhani2023microexponents,rouhani2023microscaling,ocp2023mx}.

NVIDIA demonstrated end-to-end pretraining with \nv{} by combining the
format with RHT for the weight-gradient GEMM, two-dimensional (2D) weight
scaling, SR for the back-propagated output gradient in both backward GEMMs,
and BF16 final layers~\citep{nvidia2025nvfp4}.  The Transformer Engine
(TE) recipe also computes a tensor maximum before each quantization.  These
reductions, RHT operators, and BF16 final layers increase runtime work and
implementation complexity.  We therefore ask:

\begin{quote}
\emph{Can a block-scale format with more dynamic range make FP4 training
stable with a simpler pretraining recipe?}
\end{quote}

\uefp{} combines 4-bit E2M1 payloads with 8-bit \ue{} block scales.  Because
block scales are nonnegative, \ue{} removes the unused sign bit from E4M3 and
uses five exponent bits plus three fraction bits.  Recent FP4-training studies
evaluate differentiable and mixed-precision quantization, MXFP4, native FP4
linear layers, and end-to-end NVFP4
pretraining~\citep{wang2025optimizing,zhou2025exploring,tseng2025mxfp4,
castro2025quartet,chmiel2025fp4all,nvidia2025nvfp4,panferov2026quartet2,
chen2026tetrajetv2,cook2025fouroversix}.  Hu et al.\ and Fasoli et al.\
specifically examine \ue{} as a block-scale
format~\citep{hu2025designspace,fasoli2026finer}.  We investigate the numerical
recipe required for end-to-end \uefp{} pretraining.

We make four contributions:

\begin{itemize}
  \item We introduce a \uefp{} recipe with per-operand, 50-step periodic
        sample-and-hold tensor scaling, 2D weight scaling, and SR only for the
        upstream-gradient operand in the two backward GEMMs.  The recipe omits
        RHT and uses FP4 in all 112 eligible internal linears.
  \item We train one seed-42 Nemotron-H 8B trajectory per configuration for
        188.7 billion tokens.  Across 12 checkpoints per trajectory, with
        quantization active for every FP4 path, the proposed block-16 path
        finishes below native Transformer Engine \nv{} in held-out NLL.  A
        pinned 146-task likelihood evaluation provides downstream results for
        all seven final checkpoints.
  \item We derive a range bound that links the \ue{} scale target to stale-maximum
        headroom and small-scale representation: the default target provides
        roughly $137\times$ growth headroom, while the targeted $T=2048$
        override retains $30\times$.
  \item We identify a software FP4 GEMM output model that matches the tested
        native reduction and rounding behavior in deterministic full-model
        controls.  A separate joint native \nv{} execution ablation records
        21.2\% higher model-body token throughput after disabling RHT and using
        FP4 in the 16 otherwise BF16-exempt final-block projections.
\end{itemize}

\section{Related Work}
\label{sec:related-work}

4-bit training predates native FP4 tensor cores: Sun et al.\ combined INT4
weights and activations with a radix-4 floating-point gradient format and
adaptive gradient scaling~\citep{sun2020ultralow}, while Chmiel et al.\
combined logarithmic gradient quantization with unbiased
rounding~\citep{chmiel2023accurate}.  Both studies identify gradient
range and quantization bias as central constraints on 4-bit training.

Large language model (LLM) studies subsequently tested different formats for
weights, activations, and gradients.  The initial
microscaling experiments use MXFP4 weights with MXFP6 activations and
gradients~\citep{rouhani2023microscaling}.  Wang et al.\ use simulated E2M1
weight and activation quantization with differentiable gradient estimation and
sparse outlier compensation, whereas Zhou et al.\ use different precisions
across modules and training stages~\citep{wang2025optimizing,zhou2025exploring}.  Tseng et al.\ place the backward
GEMMs in MXFP4 using SR and RHT~\citep{tseng2025mxfp4}; TetraJet studies
oscillation in MXFP4 vision-transformer training~\citep{chen2025tetrajet}; and
Metis applies spectral-domain quantization to 4-bit weights, activations, and
gradients (W4A4G4) in LLM
training~\citep{cao2026metis}.  Complementary scaling-law experiments quantify
the effects of exponent--fraction allocation, scale granularity, model size,
and token budget~\citep{sun2025scaling}.

Recent work increasingly targets FP4 across the major forward, data-gradient,
and weight-gradient GEMMs.  Quartet analyzes forward reconstruction and
backward-estimator error; \emph{FP4 All the Way} develops selective SR for
fully quantized GEMMs; and Quartet II replaces conventional SR with a
lower-error unbiased microscaling estimator~\citep{castro2025quartet,chmiel2025fp4all,panferov2026quartet2}.  NVIDIA's
long-horizon \nv{} recipe combines RHT, 2D scaling, SR, and BF16 final
layers~\citep{nvidia2025nvfp4}.  Chmiel et al.\ simulate \nv{} training of a 7B
model on 256 Intel Gaudi2 accelerators.  They use SR for the back-propagated
gradient in both backward GEMMs and additionally for the saved forward
activation used to form weight gradients; weights and forward activations
otherwise use nearest rounding~\citep{chmiel2025fp4all}.  To close the remaining loss gap,
they add a short quantization-aware fine-tuning phase that keeps the forward
pass in FP4 but runs the backward and update GEMMs in BF16.  Our recipe instead
uses \ue{} block scales, rounds the saved activation to nearest-even, applies
SR only to the back-propagated output gradient, uses periodic sample-and-hold
tensor references, and completes without a BF16 backward/update phase.
TetraJet-v2 adds oscillation suppression and
outlier control, while \emph{Four Over Six} selects between two block targets to
reduce NVFP4 quantization error~\citep{chen2026tetrajetv2,cook2025fouroversix}.  Rahimifar et al.\ instead use
transposition-invariant 2D blocks, with MXFP8 retained for sensitive query and
key projections~\citep{rahimifar2026transposition}.  A unified study by Agrusa
et al.\ compares these ingredients on dense and mixture-of-experts models at
horizons up to 1 trillion tokens and reports that its evaluated 8B \nv{}
recipes still require BF16 final layers at long
horizons~\citep{agrusa2026whatmatters}, providing a longer-horizon comparison
point for our 188.7-billion-token experiments.

Other hardware and scope extensions include native MXFP4 experiments on AMD
hardware, which isolate weight-gradient quantization as a major failure
source~\citep{cim2026mxfp4}; HiFloat4 training on Ascend neural processing
units (NPUs)~\citep{taghian2026hifloat4}; and FP4 quantization of optimizer and attention
components~\citep{ding2026fullstack}.  Within this broader literature, Hu et
al.\ identify \ue{} as a range--precision compromise for FP4 training, and
Fasoli et al.\ analyze its scale coverage using pretrained-model tensor
distributions~\citep{hu2025designspace,fasoli2026finer}.  Our study tests
whether this wider unsigned block scale permits delayed tensor scaling without
RHT or a BF16 exemption for final-block projections and extends software emulation beyond operand
quantization to the observed native \nv{} GEMM outputs.

\section{Background: FP4 Microscaling}

\subsection{Low-precision floating-point formats}

A floating-point number format stores a sign, an exponent, and a fraction
field.  Together with any implicit leading bit, the fraction field forms the
significand:
\begin{equation}
  x \approx (-1)^s \times \text{significand} \times 2^{\text{exponent}}.
\end{equation}
The notation E$e$M$m$ means $e$ exponent bits and $m$ explicit fraction bits;
a sign bit is normally separate.  More exponent bits increase the range between
very small and very large numbers.  More fraction bits place more representable
points inside that range.

Quantization maps a high-precision value to one of these representable points.
Let $Q_F(x)$ denote rounding $x$ into format $F$.  Values with magnitude above
the largest finite value can saturate or overflow, whereas nonzero values with
magnitude below the smallest positive representable value can underflow to
zero.  Saturation clips an outlier's magnitude, while underflow can remove a
small component entirely, such as an element of a weight gradient.

\subsection{Microscaling}

Microscaling keeps the payload narrow but gives each block $\mathcal B$ a scale
$s_{\mathcal B}$.  We use $B=|\mathcal B|$ for the number of payload values in
a block.  For a tensor-level multiplier $g$, a value is reconstructed as
\begin{equation}
  \widehat{x}_i = g\,\widehat{s}_{\mathcal B}\,q_i,
  \qquad
  q_i = Q_{\mathrm{E2M1}}\!\left(\frac{x_i}{g\,\widehat{s}_{\mathcal B}}\right),
  \qquad i \in \mathcal B.
  \label{eq:microscaling}
\end{equation}
Blockwise scaling lets the E2M1 codebook track each block's local magnitude
rather than the dynamic range of the full tensor.

Table~\ref{tab:formats} summarizes the relevant systems.  \mx{} is the Open
Compute Project (OCP) standard: 32 E2M1 values share an E8M0 scale.  E8M0 has a very wide exponent
range, but no fraction bits, so its scales are powers of two.  \nv{} instead
uses a block of 16 and a finer E4M3 scale.  Because E4M3 has less range, \nv{}
adds a 32-bit floating-point (FP32) scale for the whole
tensor~\citep{nvidia2026tedocs}.  The global
scale aligns the tensor with the range of the local E4M3 scales.

\begin{table}[H]
  \centering
  \scriptsize
  \caption{Summary of considered FP4 block formats.  The payload is E2M1 in
  all three FP4 systems; the systems differ in how each block is scaled.}
  \label{tab:formats}
  \begin{tabularx}{\textwidth}{@{}p{0.08\textwidth}>{\centering\arraybackslash}p{0.05\textwidth}p{0.14\textwidth}p{0.14\textwidth}p{0.22\textwidth}Y@{}}
    \toprule
    System & Block size & Scale format & Tensor-scale rule & Useful scale codes & Main tradeoff \\
    \midrule
    \mx{} & 32 & unsigned E8M0 & none & All codes are nonnegative; no fraction bits & Very wide range; coarse power-of-two steps \\
    \nv{} & 16 & signed E4M3 & FP32 global scale & Three fraction bits; negative codes are unusable for scales & Fine local spacing; narrower positive range \\
    \uefp{} & 16 & unsigned E5M3 & optional; delayed & Three fraction bits; every code is nonnegative & Fine local spacing over a much wider useful range \\
    \bottomrule
  \end{tabularx}
\end{table}

\subsection{The NVIDIA/Transformer Engine \nv{} pretraining recipe}

An outlier is much larger than the other values in its 16-value quantization
block.  It can determine the scale and leave few distinct E2M1 values
for the remaining entries.  Across blocks, the limited positive range of the
E4M3 block scale constrains the ratio of local scales under one tensor scale.
NVIDIA's \nv{} pretraining recipe uses the following techniques to manage
these effects~\citep{nvidia2025nvfp4}:

\begin{itemize}
  \item \textbf{Randomized Hadamard transform (RHT).}  The recipe applies
        matched transforms only to the saved
        activation and output-gradient operands of the weight-gradient GEMM.
        The orthogonal transforms redistribute large coordinates before
        quantization while preserving the full-precision dot product.  Similar
        rotations are widely used to suppress quantization
        outliers~\citep{ashkboos2024quarot}.
  \item \textbf{Two-dimensional weight scaling.}  A $16\times16$ weight tile
        shares scale information so that the rowwise and columnwise views used
        in forward and backward GEMMs remain consistent.
  \item \textbf{Stochastic rounding.}  The back-propagated output-gradient
        tensor is stochastically rounded when it is quantized for both backward
        GEMMs; weights and forward activations use nearest rounding.
  \item \textbf{BF16 final layers.}  The published recipe leaves the eligible
        linear projections in its final eight hybrid blocks in BF16 rather than
        quantizing those projections to FP4.
  \item \textbf{Tensor scaling.}  The maximum absolute value (\emph{amax}) of a tensor sets
        the FP32 global scale.  Current scaling requires finding that maximum
        before quantization.  Transformer Engine's documented FP8
        \texttt{DelayedScaling} records an amax at every quantization and
        derives the next scale from an amax-history window~\citep{nvidia2026delayeddocs}.
\end{itemize}

Together, these operations define the Transformer Engine \nv{} recipe used as
the baseline in this report.  We use $D$ for the number of optimizer steps
between tensor-maximum refreshes.  The evaluated baseline uses current-tensor
scaling ($D=1$); Transformer Engine's documented \texttt{DelayedScaling}
mechanism applies to FP8.

\section{Unsigned E5M3 Block Scales}

\subsection{Reallocating the unused sign bit}

A block scale is a magnitude, so it is never negative.  Signed E4M3 reserves
one of its eight bits for a sign that is always zero.  \ue{} reuses this bit as
a fifth exponent bit:
\begin{center}
  \eformat{E4M3 scale: [sign][4 exponent][3 fraction]}\\
  \eformat{UE5M3 scale:       [5 exponent][3 fraction]}
\end{center}
Hu et al.\ introduced \ue{} as an FP4 scale-format choice; Fasoli et al.\ later
analyzed its scale coverage over pretrained-model tensor
distributions~\citep{hu2025designspace,fasoli2026finer}.
At a fixed exponent, E4M3 and \ue{} have the same spacing because both keep
three fraction bits.  \ue{} does not create an extra fraction bit.  Instead,
it stops spending about half of the code space on negative values that a scale
can never use.  The result is roughly twice as many useful nonnegative scale
encodings, which increases the range of the resulting floating-point format.
In practice, \ue{} offers more usable nonnegative scale encodings over a wider
range, although its relative spacing is not finer where the two formats overlap.

For the finite format used here, the largest \ue{} value is
\begin{equation}
  U_{\max}
  = \left(1 + \frac12 + \frac14 + \frac18\right)2^{15}
  = 1.875\times 32768
  = 61{,}440.
  \label{eq:ue5m3max}
\end{equation}
Its smallest normal value is $2^{-14}$ and its smallest subnormal is $2^{-17}$.
Table~\ref{tab:scale-range} compares this with the finite E4M3 scale used by
\nv{}.  The maximum is about $137\times$ larger, and the smallest nonzero value
is $256\times$ smaller.

\begin{table}[H]
  \centering
  \caption{8-bit block-scale formats.  Both have three fraction bits, but
  \ue{} uses every code for a nonnegative scale and spends the recovered sign
  bit on range.}
  \label{tab:scale-range}
  \begin{tabular}{@{}lccccc@{}}
    \toprule
    Format & Sign bits & Exponent bits & Fraction bits & Smallest subnormal & Largest finite \\
    \midrule
    E4M3 & 1 & 4 & 3 & $2^{-9}$ & 448 \\
    \ue{} & 0 & 5 & 3 & $2^{-17}$ & 61,440 \\
    \bottomrule
  \end{tabular}
\end{table}

Under a shared tensor reference, one block may contain small values while
another contains a large outlier.  E4M3 can run out of
block-scale range, forcing the tensor-wide scale to move and making small
blocks less precise.  \ue{} can represent both block scales directly over a
much wider interval.  Fasoli et al.\ show that this can remove global scaling
for weights and activations in their setting~\citep{fasoli2026finer}.  Our
implementation instead retains a tensor reference and refreshes it once every
50 optimizer steps.

\section{Training Method}
\label{sec:training-method}

\subsection{Periodic refresh of tensor maxima}

For an operand $x$ at step $t$, define its current maximum magnitude as
\begin{equation}
  a_t = \max_i |x_{t,i}|.
\end{equation}
Current scaling uses $a_t$ immediately.  Our periodic-refresh implementation
instead samples and caches $a_t$ only on refresh steps.  Let $\tau_{t-1}$ be
the most recent refresh step before step $t$.  The reference used at step $t$
is
\begin{equation}
  \widetilde a_t =
  \begin{cases}
    a_t, & \text{if no cache exists or } t-\tau_{t-1} \ge D,\\
    a_{\tau_{t-1}}, & \text{otherwise,}
  \end{cases}
  \label{eq:delayed-amax}
\end{equation}
On a refresh step, $\tau_t=t$; otherwise $\tau_t=\tau_{t-1}$ and
$\widetilde a_t=\widetilde a_{t-1}=a_{\tau_{t-1}}$.  Thus
$\widetilde a_{t-1}$ is the cached amax from the most recent refresh, not a
$D$-step window maximum (a sample-and-hold rule).

If $r$ is a refresh step, the cache holds $a_r$ for $r\le t<r+D$ and replaces
it with the newly sampled $a_{r+D}$ at the next refresh.

This periodic-refresh rule is the implementation evaluated here.  It differs
from Transformer Engine's history-based FP8 \texttt{DelayedScaling}, which
records an amax at each iteration and uses the window maximum by default (or
the most recent entry when configured)~\citep{nvidia2026delayeddocs}.  Each quantized activation, weight, and
gradient operand has a separate cache.  We use $D=50$ with no additional
multiplicative safety margin (factor 1).  The block maximum is still computed
when that block is quantized; the tensor-wide maximum reduction and
cached-reference update occur only on refresh steps.  The caches start empty at
process launch and are not serialized in checkpoints, so a resumed training
process starts a new periodic-refresh phase rather than restoring the previous
held maxima.

Avoiding block-scale saturation requires the current tensor maximum to remain
within the cached reference's headroom.  The next subsection quantifies that
condition.

\subsection{Treating NVIDIA's fixed 448 scale as a tunable target}

The NVIDIA recipe maps the tensor reference to a fixed scale target of 448,
the largest positive finite E4M3 value.  We retain 448 as our default, but the
much wider \ue{} codebook lets us ask what happens when this target is moved.
We denote the chosen target by $T$.  It is not an input clamp and does not
change the 4-bit E2M1 payload; it only changes where the current tensor lands
inside the block-scale codebook, whose largest \ue{} value is 61,440.

The target moves the finite block-scale codebook over the real block maxima.
On a refresh step, the encoder measures a tensor maximum $g$.  For the next
$D-1$ steps it keeps that same reference, even if the tensor changes.  The
target $T$ chooses where a block as large as $g$ lands in the codebook.
Increasing $T$ slides the representable real-value window downward, helping
smaller block scales stay nonzero.  Decreasing $T$ slides the window upward,
leaving more room if the tensor grows before the next refresh.  It does not add
bits or change the spacing within an exponent band.

Let $F=6$ be the largest E2M1 payload magnitude and let $g=\widetilde a_t$ be
the cached tensor reference.  The encoder first forms the tensor multiplier
\begin{equation}
  G=\frac{TF}{g}.
\end{equation}
For a block with maximum magnitude $a_{\mathcal B}$, the ideal block-scale code before
\ue{} rounding is therefore
\begin{equation}
  u_{\mathcal B}=\frac{a_{\mathcal B}G}{F}=\frac{a_{\mathcal B}T}{g},
  \qquad
  \widehat{s}_{\mathcal B}=Q_{\mathrm{UE5M3}}(u_{\mathcal B}).
  \label{eq:scale-target}
\end{equation}
Thus, if $a_{\mathcal B}=g$, the ideal block-scale code is exactly $T$.

\paragraph{Worked example.}
Suppose the last measured tensor maximum was $g=100$ and is held fixed for 50
steps.  Table~\ref{tab:scale-target-example} shows the real block-maximum window
implied by the smallest nonzero \ue{} code, $2^{-17}$, and the largest code,
61,440.  A target of 2,048 maps the same block to a code that is
$2{,}048/448\approx4.6$ times larger than a target of 448.  This preserves
smaller scales, but the largest block that fits before saturation falls by the
same factor.

\begin{table}[H]
  \centering
  \small
  \caption{Changing $T$ slides one fixed \ue{} codebook over the real block
  maxima.  Values use $g=100$; ``growth room'' is the largest supported block
  maximum divided by the cached tensor maximum.}
  \label{tab:scale-target-example}
  \begin{tabular}{@{}lrrrr@{}}
    \toprule
    Target $T$ & Code for $a_{\mathcal B}=g$ & Smallest nonzero $a_{\mathcal B}$ & Largest $a_{\mathcal B}$ & Growth room \\
    \midrule
    448   & 448   & $1.7\times10^{-6}$ & 13,714 & $137\times$ \\
    2,048 & 2,048 & $3.7\times10^{-7}$ & 3,000  & $30\times$ \\
    \bottomrule
  \end{tabular}
\end{table}

\paragraph{Proposition 1 (unsaturated delayed-scale range).}
Let $U$ be the largest finite block-scale code.  Under
Equation~\ref{eq:scale-target}, block $\mathcal B$ avoids scale saturation exactly when
\begin{equation}
  a_{\mathcal B} \le \frac{Ug}{T}.
  \label{eq:effective-range}
\end{equation}
If $a_t$ is the current tensor maximum and
$\rho_t=a_t/g$ measures how stale the cache has become, then every block is
unsaturated whenever
\begin{equation}
  T\rho_t\le U,
  \qquad\text{or equivalently}\qquad
  T\le\frac{U}{\rho_t}.
  \label{eq:stale-headroom}
\end{equation}
The proof is one substitution: saturation starts at $u_{\mathcal B}>U$, and
$u_{\mathcal B}=a_{\mathcal B}T/g$.

This proposition formalizes the window in
Table~\ref{tab:scale-target-example}.  A smaller $T$ leaves more overflow
headroom when the current tensor grows above its cached reference.  A larger
$T$ keeps smaller block scales away from zero but leaves less room for growth.
The delay $D$ and target $T$ must therefore be chosen together: a longer delay
allows a larger stale ratio $\rho_t$, which generally favors a smaller target.

For E4M3, NVIDIA's fixed choice gives $U=T=448$.  Under this mapping, a tensor
already at its cached maximum has no additional stale-growth headroom.  With
\ue{}, $U=61{,}440$ while the default target remains 448.  At $T=448$,
Equation~\ref{eq:stale-headroom} allows a stale ratio up to roughly
$137\times$ before block-scale saturation.  Raising the target to 2,048 spends
some of that headroom to map small block scales about $4.6\times$ farther from
zero.
The extended range can be allocated between stale-maximum headroom and the
representation of smaller block scales without changing the E2M1 payload.

For the 8B model, the proposed recipe uses $T=448$ by default for all FP4
operands.  We raise it to $T=2048$ only for $dY$ in the weight-gradient GEMMs
of the four final MLP \texttt{mixer.down\_proj} modules (zero-based layers 45,
47, 49, and 51).  This moves their block-scale codes farther from zero,
increasing the underflow margin for small gradient components while retaining
$61{,}440/2{,}048=30\times$ stale-growth headroom.

\begin{figure}[H]
  \centering
  \includegraphics[width=\textwidth]{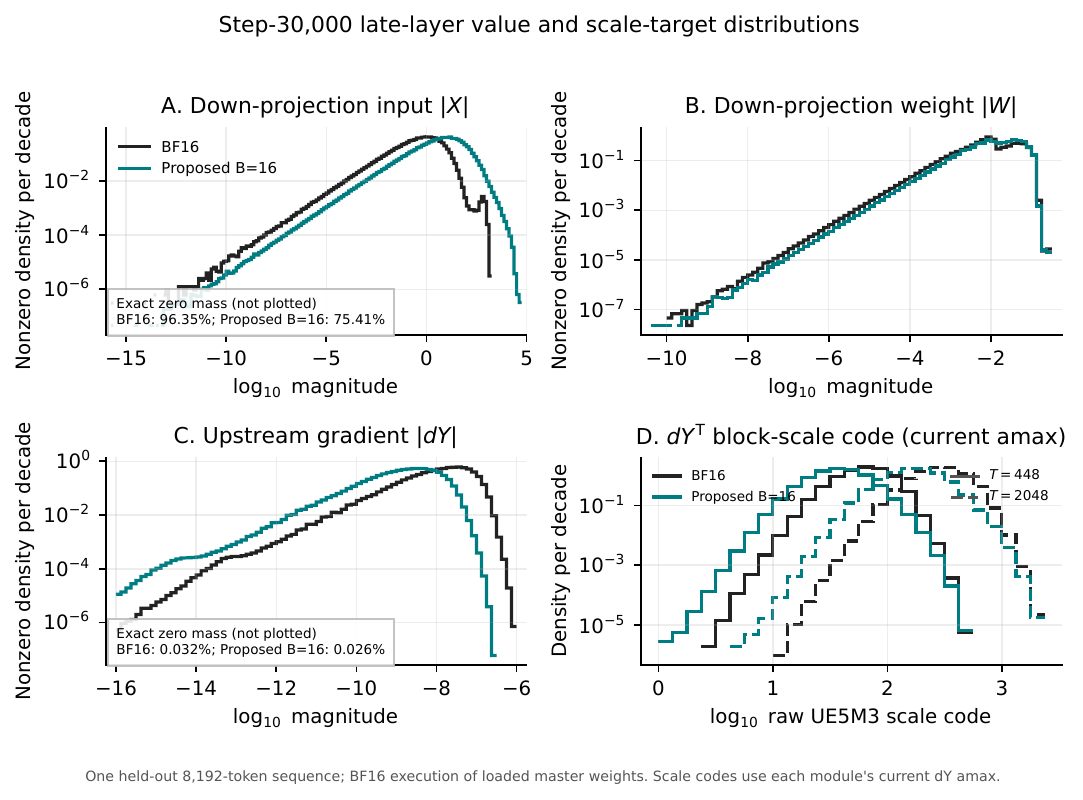}
  \caption{Step-30,000 checkpoint snapshot for the four final MLP
  \texttt{mixer.down\_proj} modules.  Absolute-value histograms pool the saved
  activation $X$, stored BF16 master weight $W$, and upstream gradient $dY$
  over one fixed held-out sequence.  The scale-code distributions map block-16
  maxima of $dY^{\mathsf T}$ in its weight-gradient GEMM layout to ideal
  pre-rounding \ue{} scale codes under $T=448$ and $T=2048$.}
  \label{fig:scale-target-checkpoint-snapshot}
\end{figure}

\paragraph{Checkpoint snapshot.}
Figure~\ref{fig:scale-target-checkpoint-snapshot} compares the step-30,000 BF16
reference and proposed block-16 checkpoints.  We load their stored BF16 master
weights and run one fixed, zero-dropout BF16 training-mode forward/backward pass
on the first 8,192-token sequence in the held-out order, without an optimizer
update.  Pooled over the four modules, the saved activation $X$ is 96.35\% zero
for the BF16 checkpoint and 75.41\% zero for the proposed checkpoint, with
maxima of 1,408 and 48,384, respectively.  The corresponding $dY$ tensors are
nearly dense (about 0.03\% zeros in both), while the weights and block maxima
formed from $dY^{\mathsf T}$ contain no zeros.

Using each module's current $dY$ amax as $g$, neither target produces a scale
code that rounds to zero before repair or a saturated scale code in this
snapshot.  It therefore illustrates codebook placement rather than an observed
underflow or saturation event.  This is one held-out checkpoint snapshot, not
a training-wide distribution or a replay of the $D=50$ cache history.  The
$dY$ histogram and its block maxima describe training-only quantities and have
no inference analogue.

\paragraph{Matched smaller-model control.}
An earlier 350M experiment provides a concrete training example.  Both runs
used the same source revision, seed 42, 10,000 updates, and $D=50$.  The control
used $T=448$ throughout; the comparison changed only the $dY$ target to
$T=2048$ in the \texttt{feed\_forward.w2} weight-gradient GEMMs of the final
four layers.  Over the final 250 updates, the mean of the ten logged losses was
2.90904 at $T=448$ and 2.90248 with the targeted $T=2048$ override, a
difference of $-0.00656$ (Table~\ref{tab:scale-target-350m-control}).

\begin{table}[H]
  \centering
  \small
  \caption{Matched 350M scale-target control.  The final-window mean is the
  arithmetic mean of ten global-average training-loss values logged every 25
  updates from steps 9,775 through 10,000; the step-10,000 column is the final
  logged value.  Values are rounded to five decimal places, and each row is one
  seed-42 trajectory.}
  \label{tab:scale-target-350m-control}
  \begin{tabular}{@{}lrr@{}}
    \toprule
    Wgrad-$dY$ target in final four \texttt{feed\_forward.w2} layers
      & \shortstack{Final-window mean\\training loss}
      & \shortstack{Step-10,000\\training loss} \\
    \midrule
    $T=448$ (default) & 2.90904 & 2.86274 \\
    $T=2048$          & 2.90248 & 2.85795 \\
    \bottomrule
  \end{tabular}
\end{table}

This smaller-model control motivated the targeted override.  The 8B experiment
evaluates the complete recipe rather than an isolated scale-target ablation.

\subsection{Stochastic rounding preserves small gradients in expectation}

Suppose $z$ lies between adjacent representable values $q^-$ and $q^+$.  SR
chooses
\begin{equation}
  Q_{\mathrm{SR}}(z)=
  \begin{cases}
    q^-, & \text{with probability } \dfrac{q^+-z}{q^+-q^-},\\[6pt]
    q^+, & \text{with probability } \dfrac{z-q^-}{q^+-q^-}.
  \end{cases}
  \label{eq:sr}
\end{equation}
Then $\mathbb{E}[Q_{\mathrm{SR}}(z)]=z$.  This construction is unbiased.  Each
draw introduces zero-mean rounding error instead of systematically removing
values below a deterministic rounding threshold.  Unbiased SR is widely used
in low-precision training~\citep{gupta2015limited,ozkara2025sr}.  Practical
implementations must also handle randomness carefully: natural constructions
that use only a few random bits can introduce bias even though the ideal rule
in Equation~\ref{eq:sr} is unbiased~\citep{fitzgibbon2025fewbits}.

For a linear $Y=XW^{\mathsf T}$, the two backward GEMMs are
$dX=dY\,W$ and $dW=dY^{\mathsf T}X$.  We use SR only for the E2M1 payload of
the upstream-gradient operand $dY$ in both GEMMs.  The saved activation $X$,
weight $W$, forward operands, and all block-scale codes use deterministic
round-to-nearest, ties-to-even.  Transformer Engine likewise applies SR to the
$dY$ operand in these two backward GEMMs~\citep{nvidia2025nvfp4,nvidia2026tedocs};
the primary recipes differ in their scale format and lifecycle, RHT, and
final-block precision rather than in which backward operand receives SR.

\subsection{Recipe comparison}

Table~\ref{tab:primary-recipes} compares the format, scale lifecycle, rounding,
layer coverage, and GEMM path of the two primary recipes.  ``All eligible''
means the converted feed-forward-network (FFN), attention, and Mamba-block
input/output linear projections.  The output projection is outside this set
and evaluated in FP32 in both recipes.

\begin{table}[H]
  \centering
  \small
  \caption{Detailed comparison of the Transformer Engine
  \nv{} recipe and our proposed \uefp{} recipe.}
  \label{tab:primary-recipes}
  \begin{tabularx}{\textwidth}{@{}lYY@{}}
    \toprule
    Component & Transformer Engine \nv{} recipe & Our proposed \uefp{} recipe \\
    \midrule
    Payload / block scale
      & E2M1 payload, signed E4M3 scale, blocks of 16
      & E2M1 payload, unsigned E5M3 scale, blocks of 16 \\
    Tensor reference
      & Current amax ($D=1$)
      & Separate cached activation, weight, and gradient maxima, refreshed
        every $D=50$ steps \\
    Scale target $T$
      & Fixed at 448
      & $T=448$ by default; $T=2048$ only for Wgrad-$dY$ in the four final MLP
        \texttt{mixer.down\_proj} modules (zero-based layers 45, 47, 49, and 51) \\
    RHT
      & Saved-activation and output-gradient operands of the weight-gradient
        GEMM only
      & None \\
    Weight scaling
      & 2D, aligned with both GEMM views
      & 2D, aligned with both GEMM views \\
    Deterministic rounding
      & Transformer Engine nearest rounding on forward operands
      & Ties-to-even on activations, weights, and block scales; a block scale
        that rounds to zero is replaced with 1 \\
    Stochastic rounding
      & Upstream-gradient ($dY$) operand in $dX=dY\,W$ and
        $dW=dY^{\mathsf T}X$
      & Upstream-gradient ($dY$) operand in $dX=dY\,W$ and
        $dW=dY^{\mathsf T}X$ \\
    Eligible-linear coverage
      & 96 use FP4; 16 projections in the final eight hybrid blocks remain
        BF16; output head in FP32
      & All 112 use FP4; no BF16 exemption; output head in FP32 \\
    Matrix-multiply path
      & Native Blackwell FP4 through Transformer Engine
      & Custom quantization plus a software FP4 GEMM model matched to
        native-output probes, described in Section~\ref{sec:gemm-output-model} \\
    \bottomrule
  \end{tabularx}
\end{table}

The proposed run uses a deterministic zero-scale rule: after \ue{} rounding,
any zero block-scale code is replaced with 1 before reciprocal computation and
dequantization.  This zero-scale replacement rule is separate from SR, which randomly chooses
neighboring payload values to preserve their expectation.

Table~\ref{tab:control-recipes} adds two controls for format--recipe
interactions.

\begin{table}[H]
  \centering
  \scriptsize
  \caption{Experimental recipe matrix.  ``Native'' means
  Transformer Engine's hardware FP4 path; ``probe-matched'' means a software
  FP4 GEMM model fitted to native-output probes and specified in
  Section~\ref{sec:gemm-output-model}.  The table reports the effective
  tensor-scale refresh interval in optimizer steps.}
  \label{tab:control-recipes}
  \begin{tabularx}{\textwidth}{@{}>{\raggedright\arraybackslash}p{0.22\textwidth}cccccY@{}}
    \toprule
    Recipe & Block scale & $D$ & RHT & Final blocks with BF16 linears & GEMM & Purpose \\
    \midrule
    Transformer Engine \nv{} recipe & E4M3 & 1 & yes & 8 & native & Published baseline \\
    Proposed \uefp{} recipe & \ue{} & 50 & no & 0 & probe-matched & Primary experiment \\
    Native \nv{} no-RHT/all-linears ablation & E4M3 & 1 & no & 0 & native & Native execution ablation \\
    \uefp{} with Transformer Engine settings & \ue{} & 1 & yes & 8 & probe-matched & Transformer Engine settings with \ue{} block scales \\
    \bottomrule
  \end{tabularx}
\end{table}

The proposed recipe applies the $T=2048$ override only to the Wgrad-$dY$
operands in these four final MLP modules.  Its default block size is 16; the
block-size control changes only this value to 32 while retaining the same seed,
data, optimizer, delayed-scaling controls, upstream-gradient SR, layer coverage,
GEMM emulator, and 30,000-step schedule.

\section{Modeling Observed Native \texorpdfstring{\nv{}}{NVFP4} GEMM Outputs}
\label{sec:gemm-output-model}

\subsection{Accumulation-path sensitivity}

A dot product computes
\begin{equation}
  y = \sum_{k=1}^{K} a_k b_k.
\end{equation}
In exact arithmetic, the order of addition does not matter.  In floating-point
arithmetic, each intermediate sum is rounded, so changing the reduction order
or rounding mode can change the answer.  This means two systems can decode
identical FP4 values and scales but still produce different GEMM outputs.

Our decoded-operand control reconstructs the FP4 operands as FP32 tensors and
applies a standard Torch matrix multiplication before returning the result to
BF16.  Here, ``FP32 tensors'' describes operand storage; the matrix
multiplication can still use the accelerated arithmetic selected by the
Torch/CUDA runtime.  This
decoded-operand path can group products differently and can round at different moments
from an FP4 tensor core.  In the first
sensitive example, native Transformer Engine returned 0.1630859375 while the
software path returned 0.162109375: one BF16 step apart.  Although the
discrepancy is only one BF16 step, it enters backpropagation and can affect
later updates.

\subsection{The observed reduction and product lattice}

Targeted probes of the tested Blackwell/Transformer Engine path support an
output model with three rules:

\begin{enumerate}
  \item split the dot product into groups of 64 products;
  \item make one FP32 partial sum per group, rounded to nearest-even;
  \item add the group totals in physical order, choosing the FP32 value toward
        zero whenever an addition is inexact.
\end{enumerate}

These empirical rules describe the outputs of the evaluated hardware and
software stack.  In equations, let
\begin{equation}
  p_j = \operatorname{RNE}_{32}
  \left(\sum_{k=64j}^{64j+63} a_kb_k\right),
  \label{eq:issue-partial}
\end{equation}
where $\operatorname{RNE}_{32}$ denotes FP32 round-to-nearest, ties-to-even.
We write $\operatorname{RTZ}_{32}$ for FP32 round-toward-zero.  Combine the partials in physical order using round-toward-zero on every
inexact cross-group addition:
\begin{equation}
  c_0=0,
  \qquad
  c_{j+1}=\operatorname{RTZ}_{32}(c_j+p_j).
  \label{eq:issue-rz}
\end{equation}
The decisive permutation witness is built from decoded products that are
already exact multiples of $1/1024$.  Our emulator therefore optionally
canonicalizes the unscaled result on that product lattice with
round-to-nearest, ties-to-even, and only then applies the scale product $\alpha$:
\begin{equation}
  y = \alpha\,
  \frac{\operatorname{RNE}(1024c_J)}{1024}.
  \label{eq:final-grid}
\end{equation}

The emulator implements Equation~\ref{eq:final-grid} as
\texttt{torch.round(1024 * c) / 1024}.  Thus an exact half-grid case selects
the grid point whose integer index is even; it is not rounded toward zero.
The multiplication and division by 1,024 are power-of-two shifts, and the
encoded tensor-scale product $\alpha$ is applied after this operation.
No snap, $1/1024$, and finer grids agree on all 258 permutation witnesses.
We use ties-to-even $1/1024$ canonicalization as the emulator's explicit
output rule.

Round-toward-zero does not mean that every result is rounded downward.  A
positive intermediate moves down, while a negative intermediate moves up, so
both move slightly closer to zero.  This sign-dependent rule is selected solely
by native-output parity.

The implementation uses a Triton GEMM with one BF16 dot product per 64-wide
slice, FP32 partial sums, round-toward-zero additions, and optional
product-lattice canonicalization.  We call this numerical model the
\emph{probe-matched FP4 emulator}.

\subsection{Final-grid granularity and identifiability}

We vary the final-grid denominator $d$.  For $d>0$, the emulator computes
\begin{equation}
  Q_d(c)=\frac{\operatorname{RNE}(dc)}{d},
  \qquad
  |Q_d(c)-c|\leq \frac{1}{2d}.
  \label{eq:grid-denominator}
\end{equation}
A smaller denominator means a coarser grid: it has a larger worst-case error
and maps the wider interval $[-1/(2d),+1/(2d)]$ to zero.  A larger denominator
means a finer grid.  Setting $d=0$ disables the operation.  We use powers of two
so multiplication and division by $d$ are exact binary exponent shifts before
the rounding step.

Table~\ref{tab:grid-denominator-native} applies six choices to the same 258
native permutation labels used to identify the cross-group rule.  Coarser
grids destroy native bins: $1/512$ loses all 37 occurrences of odd bin $-3371$,
and $1/256$ merges additional bins.  In contrast, $1/1024$, both finer grids,
and no final snap are identical on this corpus because its decoded products
already lie on the $1/1024$ lattice.

\begin{table}[H]
  \centering
  \small
  \caption{Final-grid denominator sweep on 258 native FP32 permutation
  witnesses.  ``None'' means $d=0$.}
  \label{tab:grid-denominator-native}
  \begin{tabular}{@{}lrrrrrr@{}}
    \toprule
    Grid & None & $1/256$ & $1/512$ & $1/1024$ & $1/2048$ & $1/4096$ \\
    \midrule
    Exact native bins & 258 & 180 & 221 & 258 & 258 & 258 \\
    Match rate (\%) & 100.0 & 69.8 & 85.7 & 100.0 & 100.0 & 100.0 \\
    \bottomrule
  \end{tabular}
\end{table}

Among the tested denominators, $d=1024$ is the coarsest canonicalization that
preserves all 258 native matches, so we use it in the emulator.

\subsection{End-to-end parity test}

We compared native Transformer Engine \nv{} against the custom quantizer plus
the probe-matched emulator in the paper's 1.2B model configuration.  This
configuration has 20 layers, width 2,048, vocabulary size 131,072, and exactly
1,291,929,600 trainable scalar parameters; ``1.2B'' is a rounded label.
Both paths started from the same BF16 parameter tensors and used the same
synthetic input and target (batch size 1, sequence length 64, model seed 1,234,
and batch seed 5,678).  Each path quantized the same 96 linear modules, kept the
final four transformer layers in BF16, and kept the output projection in FP32.
RHT and 2D weight scaling were enabled.  SR was disabled so repeated values
could be compared exactly rather than statistically.

The custom path used the probe-matched emulator in the forward GEMM, the
data-gradient GEMM that propagates error to the preceding layer, and the
weight-gradient GEMM that forms each parameter update direction.  We then ran
one cross-entropy forward and backward pass.  The result was:

\begin{itemize}
  \item logits exactly equal;
  \item loss difference exactly zero;
  \item every named parameter had a gradient in both paths, with no missing
        tensors;
  \item the compared gradient count was 1,291,929,600, exactly equal to the
        model's parameter count; and
  \item gradient relative $L_2$ error and maximum absolute error both zero.
\end{itemize}

The 1.29B comparison covers the complete parameter-gradient vector, including
gradients for quantized modules and the high-precision embedding, BF16 final layers,
and output modules.  The global checker casts each BF16 gradient losslessly to
FP32 before subtraction, establishing equality of every represented BF16
value.
Appendix~\ref{app:100step-parity} extends this check through 100 complete
AdamW updates and compares raw tensor storage rather than only represented
gradient values.  Appendix~\ref{app:gemm-ablations} gives the witness
construction, group-width and rounding ablations, broader sampled-output
comparisons, and final-grid identifiability results used to select the
emulator.

\section{Experiments}

\subsection{Setup}

We use the Nemotron-H 8B configuration, a hybrid architecture in which most
self-attention layers are replaced by Mamba layers~\citep{nvidia2025nemotronh}.
We match the disclosed architecture, sequence length, global batch, optimizer,
and schedule from Appendix A.2 of the \nv{} paper, then substitute the data
blend and train for 188.7 billion tokens.  Table~\ref{tab:setup} records each
match and substitution.

NVIDIA reports a 1T-token, two-phase blend, but the underlying 8B examples and
mixture weights are unavailable.  We instead use a fixed OLMo-family mixture
containing 82\% DataComp-LM and 18\% other OLMo data.

\begin{table}[H]
  \centering
  \scriptsize
  \caption{Comparison with NVIDIA's 8B setup.  ``Exact'' denotes a
  field-by-field match; ``Partial'' means that only the listed fields match;
  ``Shortened'' and ``Substituted'' identify the stated horizon and data
  changes.}
  \label{tab:setup}
  \begin{tabularx}{\textwidth}{@{}p{0.18\textwidth}YYp{0.12\textwidth}@{}}
    \toprule
    Component & NVIDIA 8B setup & This study & Match \\
    \midrule
    Hybrid blocks
      & 52 total: 4 attention, 24 FFN, 24 Mamba-2
      & Same 52-block arrangement
      & Exact \\
    Core widths
      & Hidden 4,096; FFN 21,504
      & Hidden 4,096; FFN 21,504
      & Exact \\
    Attention
      & 32 query heads; 4 key/value heads
      & 32 query heads; 4 key/value heads
      & Exact \\
    Mamba-2
      & 8 groups; state 128; head 64; expansion 2; convolution width 4
      & Same five dimensions
      & Exact \\
    Sequence / batch
      & 8,192 tokens; global batch 768
      & 8,192 tokens; global batch 768
      & Exact \\
    Optimizer
      & AdamW, $\beta_1=0.9$, $\beta_2=0.95$, weight decay 0.1
      & Same optimizer and coefficients
      & Exact \\
    Learning rate
      & Warmup--Stable--Decay (WSD): $8\times10^{-4}$, decaying to $8\times10^{-6}$ over the final 15\%
      & Constant through 85\%, then linear decay to 1\% of peak
      & Exact \\
    Training horizon
      & 1 trillion tokens
      & 30,000 steps, or
        $30{,}000\times768\times8{,}192=188.7$ billion tokens
      & Shortened \\
    Training data
      & Two blend phases; exact 8B examples and mixture unavailable
      & Fixed internal 82/18 OLMo-family mixture
      & Substituted \\
    Precision baselines
      & BF16; \nv{} linears with eligible projections in the final eight blocks in BF16;
        output projection in FP32
      & Same BF16 and Transformer Engine \nv{} precision placement
      & Exact \\
    Execution controls
      & Blackwell FP4; world size, seed, and clipping not disclosed
      & 32 GB200 ranks; seed 42; global-norm clipping at 1.0
      & Partial \\
    \bottomrule
  \end{tabularx}
\end{table}

The architecture, optimizer, schedule, sequence and batch sizes, and precision
placement---including the FP32 output projection---match the NVIDIA 8B setup.
The deliberate differences are the shorter training horizon and substituted
data mixture; world size, seed, and clipping are reported as study-specific
execution controls because the NVIDIA values are not disclosed.

All seven trajectories reach 30,000 optimizer steps.  The Transformer Engine
\nv{} trajectory was resumed at step 15,000 with its model, optimizer, and
schedule state restored; its data-order and random-number streams restarted
from seed 42.

Each configuration has one seed-42 trajectory, so the reported loss and
downstream-score differences are descriptive comparisons rather than estimates
of seed-to-seed variability or statistical significance.

\subsection{Inference with delayed scaling}
\label{sec:post-load-inference}

Delayed scaling introduces state during pretraining: the scale used at a given
step depends on a maximum measured earlier.  That state is an execution detail,
not part of the learned model, so it is not available when a checkpoint is
loaded for inference.  We therefore initialize inference scales explicitly
rather than attempting to reconstruct the final training cache.

Our main \uefp{} evaluations preserve the same design principle used in
pretraining.  We keep each weight tensor's amax reference fixed because weights
do not change during inference, while activation references refresh every 50 inference batches
and remain fixed between refreshes.  We start from an empty cache and use one
fixed evaluation order, making this stateful policy deterministic and
comparable across checkpoints.  Table~\ref{tab:inference-scale-summary} lists
this policy and two stateless sensitivity checks; calibration data are
disjoint from validation.

Every \uefp{} result includes FP4 fake quantization of the eligible weights and
activations; the native \nv{} results execute those operations in Transformer
Engine, and the BF16 control remains unquantized.  Thus the evaluation measures
FP4 weight and activation error under an explicitly defined inference-scale
policy.

\begin{table}[H]
  \centering
  \small
  \caption{Activation-scale choices used for quantized inference.}
  \label{tab:inference-scale-summary}
  \begin{tabularx}{\textwidth}{@{}lYY@{}}
    \toprule
    Policy & Scale rule & Role in this study \\
    \midrule
    Delayed, $D=50$ & Refresh every 50 batches and hold between refreshes
      & Main evaluation for the proposed recipe \\
    Current, $D=1$ & Recompute for every batch
      & Transformer Engine setting and sensitivity check \\
    Calibrated frozen & Estimate on separate calibration data, then hold fixed
      & Stateless sensitivity check \\
    \bottomrule
  \end{tabularx}
\end{table}

\subsection{Pretraining quality and stability}

\begin{figure}[H]
  \centering
  \includegraphics[width=0.90\textwidth]{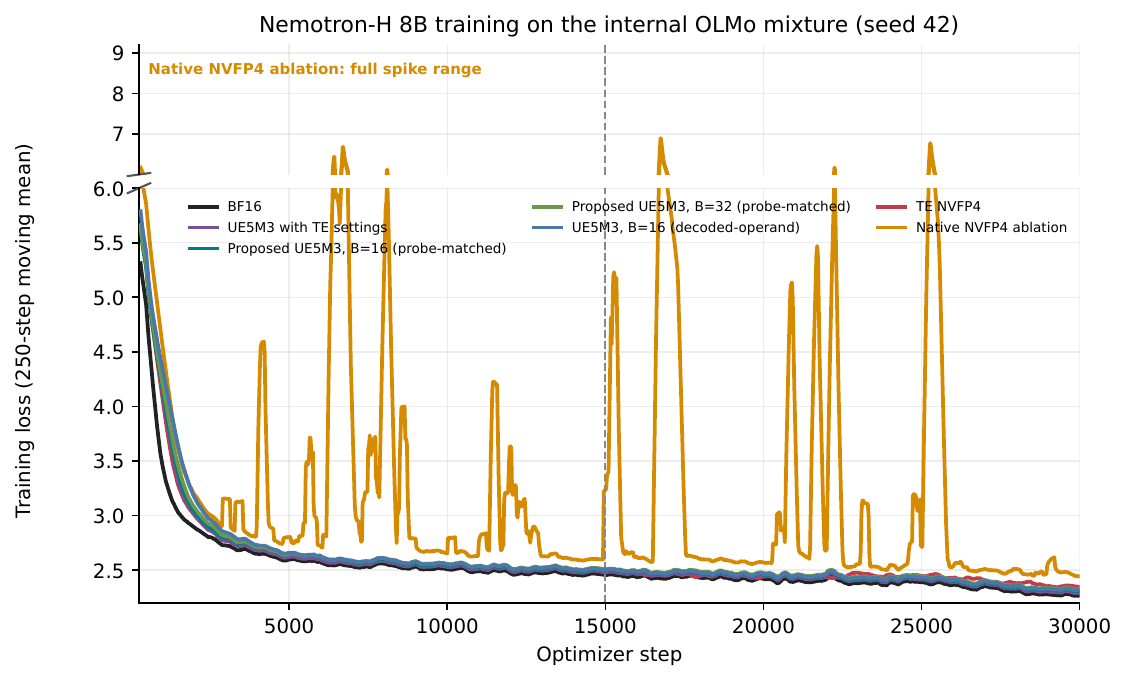}
  \caption{Training loss for the completed 8B runs.  Lower is better.  The
  split vertical axis keeps the main trajectories readable while showing the
  full native-\nv{} ablation spike range.  The dashed line marks the Transformer Engine
  optimizer resume.  Curves average 10 logged values, or 250 optimizer steps.
  $B$ is the number of values in each microscaling block; ``probe-matched'' denotes
  the probe-matched GEMM output model.}
  \label{fig:curves}
\end{figure}

Six trajectories descend smoothly after early training; the native \nv{}
no-RHT/all-linears ablation exhibits repeated loss spikes.  Figure~\ref{fig:curves}
shows the complete trajectories.

\begin{figure}[H]
  \centering
  \includegraphics[width=\textwidth]{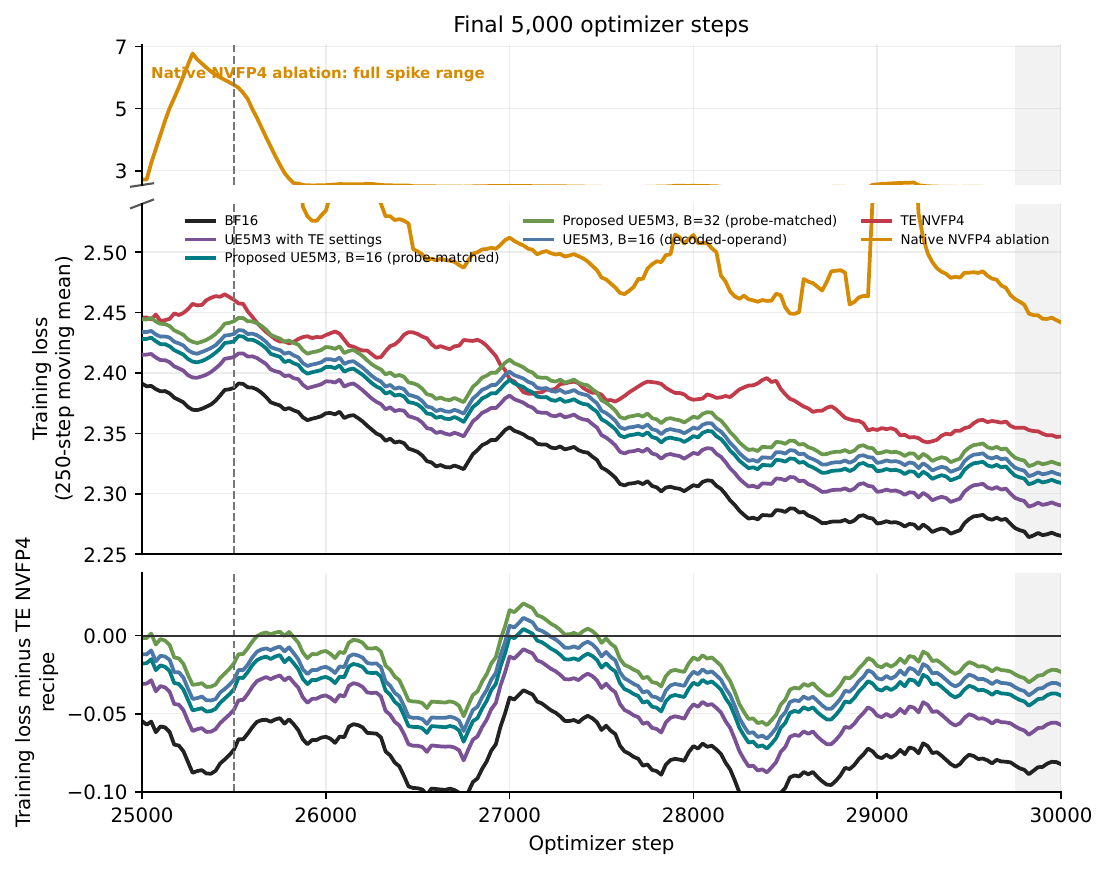}
  \caption{The final 5,000 optimizer steps.  Curves use the same 250-step
  moving mean as Figure~\ref{fig:curves}.  The split upper panels include the
  full native-\nv{} ablation trajectory.  The dashed line marks the start of linear
  decay; the gray band is the final reporting window.  In the bottom panel,
  negative values indicate lower loss than the Transformer Engine \nv{} recipe.}
  \label{fig:late-zoom}
\end{figure}

During the final 5,000 steps, all four \uefp{} trajectories have lower
final-window means and endpoint losses than the Transformer Engine \nv{}
trajectory (Figure~\ref{fig:late-zoom}).

\begin{figure}[H]
  \centering
  \includegraphics[width=0.96\textwidth]{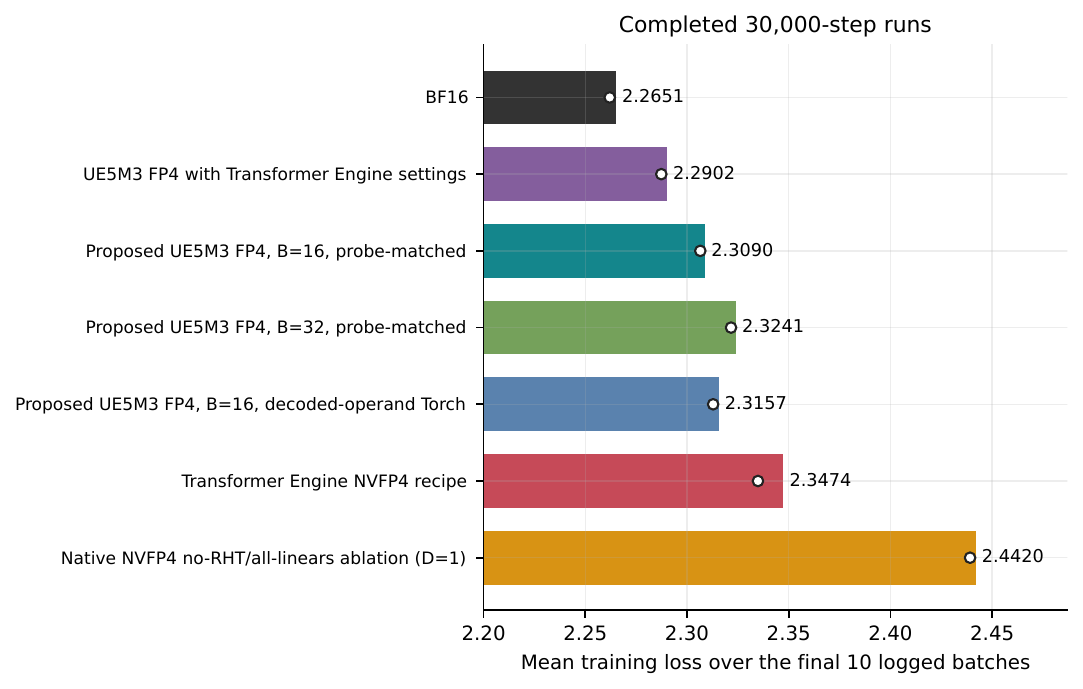}
  \caption{Means of the ten logged training losses in the final 250-step
  window.  The white marker is the exact endpoint training loss.}
  \label{fig:final-loss}
\end{figure}

Table~\ref{tab:results} and Figure~\ref{fig:final-loss} summarize the final
250-step window and report the endpoint separately.

\begin{table}[H]
  \centering
  \small
  \caption{Completed 30,000-step pretraining runs.  Lower training loss is
  better.  ``Window mean'' is the arithmetic mean of ten logged training
  losses with steps in $(29{,}750,30{,}000]$; gradient norm is the pre-clipping
  global norm.}
  \label{tab:results}
  \begin{tabularx}{\textwidth}{@{}Yrrr@{}}
    \toprule
    Run & Window-mean training loss & Endpoint training loss & Endpoint gradient norm \\
    \midrule
    BF16 & \textbf{2.2651} & 2.2620 & 0.0544 \\
    \uefp{} with Transformer Engine settings & 2.2902 & 2.2874 & 0.0352 \\
    Proposed \uefp{}, $B=16$, probe-matched & 2.3090 & 2.3065 & 0.0417 \\
    Proposed \uefp{}, $B=32$, probe-matched & 2.3241 & 2.3216 & 0.0374 \\
    Proposed \uefp{}, $B=16$, decoded-operand Torch & 2.3157 & 2.3128 & 0.0396 \\
    Transformer Engine \nv{} recipe & 2.3474 & 2.3349 & 0.0349 \\
    Native \nv{} no-RHT/all-linears ablation ($D=1$) & 2.4420 & 2.4391 & 0.4551 \\
    \bottomrule
  \end{tabularx}
\end{table}

The lowest FP4 final-window loss is 2.2902 for \uefp{} with Transformer Engine
settings, compared with 2.3474 for native Transformer Engine \nv{}.  Our
proposed block-16 recipe reaches 2.3090 while omitting RHT and applying FP4 to
all 112 eligible internal linears.  Its matched block-32 control reaches
2.3241, and its decoded-operand GEMM control reaches 2.3157.  Within these
single trajectories, the comparisons favor block size 16 and the probe-matched
GEMM output model for the proposed recipe.

\begin{figure}[H]
  \centering
  \includegraphics[width=\textwidth]{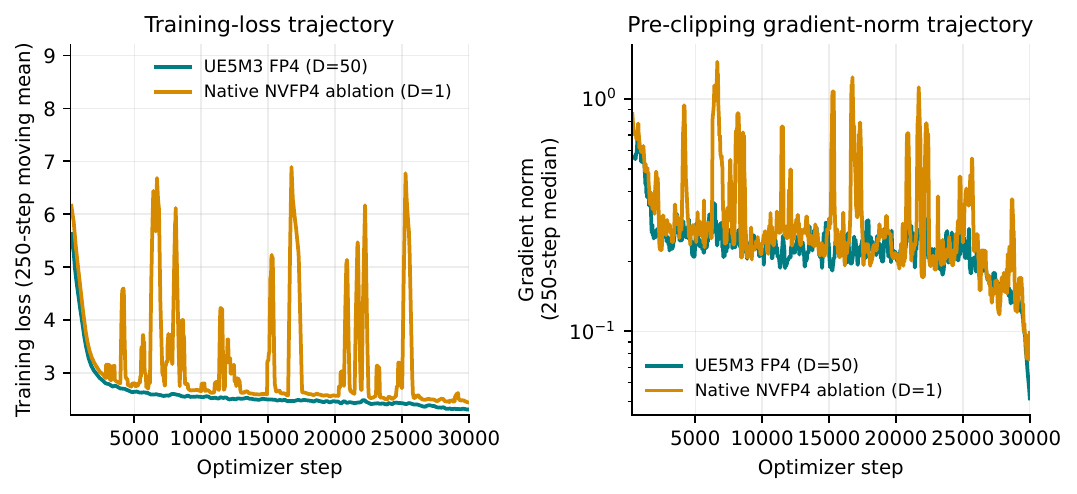}
  \caption{Training stability when all 112 eligible internal linears use FP4.
  The proposed \uefp{} recipe uses periodic scaling with $D=50$; the native
  \nv{} ablation uses current-tensor scaling and otherwise removes RHT and the
  final-block BF16 exemption.  Loss uses a 250-step moving mean and gradient
  norm a 250-step moving median.}
  \label{fig:matched}
\end{figure}

\paragraph{Stability with all eligible linears in FP4.}
Figure~\ref{fig:matched} compares the proposed \uefp{} recipe with a native
\nv{} ablation that also removes RHT and the BF16 exemption for the final-block
projections.  This is a comparison of complete recipes rather than an isolation
of the scale format: the two runs also differ in scale lifecycle and GEMM
numerics.  After step 2,500, the proposed trajectory records no logged loss
above 3 and no pre-clipping gradient norm above 1.  The native \nv{} ablation
records 205 losses above 3 and 89 gradient norms above 1, and finishes with a
final-window loss of 2.4420 versus 2.3090.

\begin{figure}[H]
  \centering
  \includegraphics[width=0.94\textwidth]{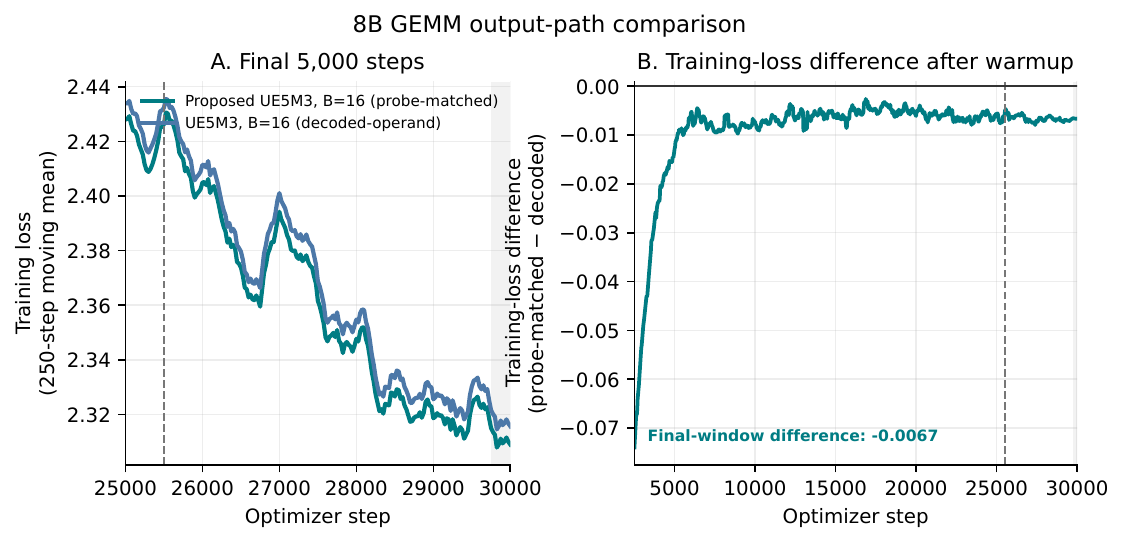}
  \caption{Matched block-16 \uefp{} runs that differ only in their software
  GEMM output model.  (A) Final 5,000 steps.  (B) Probe-matched loss minus the
  decoded-operand control after warm-up; negative values favor the
  probe-matched emulator.}
  \label{fig:gemm-training-control}
\end{figure}

\paragraph{Effect of the GEMM output model.}
A paired block-16 control changes only how the software GEMM models native FP4
reduction and output rounding.  The probe-matched emulator reaches a
final-window loss of 2.3090, compared with 2.3157 for a standard matrix
multiplication on reconstructed operands; the endpoint has the same ordering.
Figure~\ref{fig:gemm-training-control} shows that this small difference persists
through the final decay window.  Section~\ref{sec:gemm-output-model} and
Appendix~\ref{app:gemm-ablations} define the two numerical models.

\subsection{Quantized held-out validation across checkpoints}
\label{sec:quantized-validation}

We evaluate 12 checkpoints from steps 2,500 through 30,000 for all seven
trajectories on the same fixed validation stream of 768 ordered 8,192-token
sequences (6,291,456 tokens), using the inference paths in
Table~\ref{tab:quantized-validation-final}.

\begin{figure}[H]
  \centering
  \includegraphics[width=\textwidth]{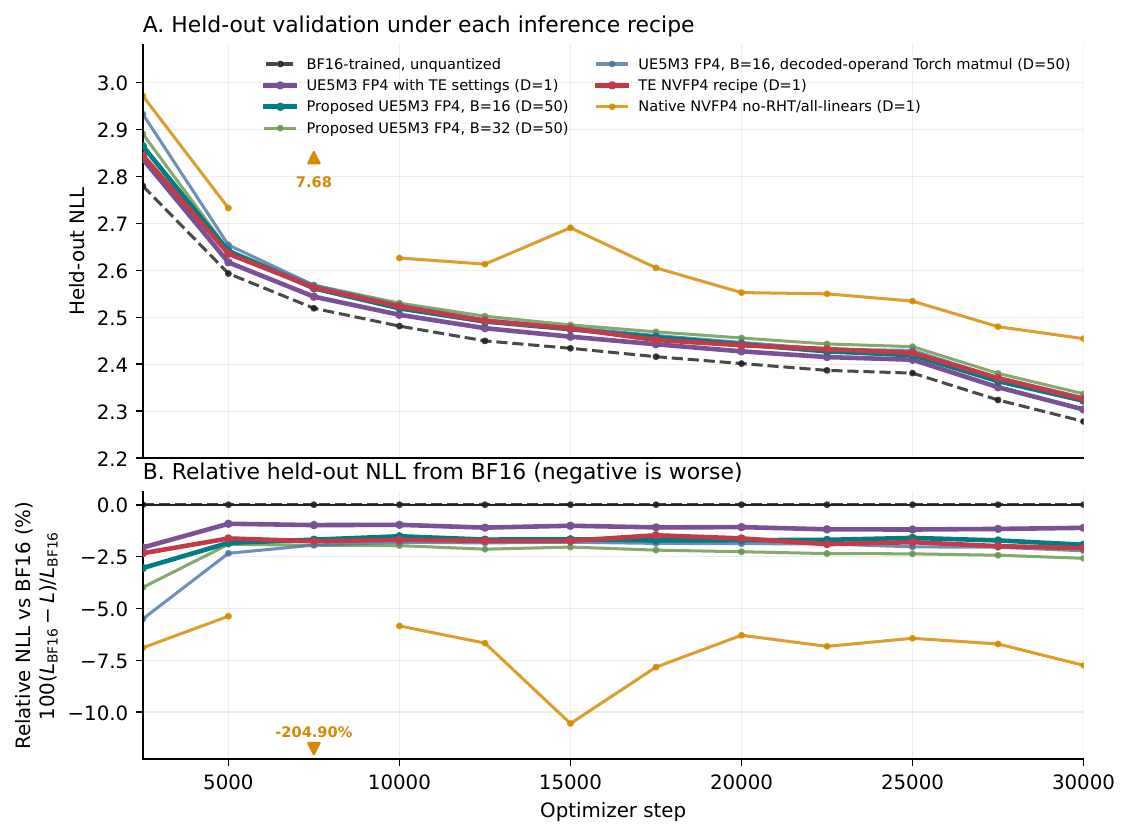}
  \caption{Held-out NLL under quantized inference at 12 checkpoints; BF16 is
  unquantized.  Panel A shows absolute NLL.  Panel B follows
  \citet{nvidia2025nvfp4} and reports
  $100(L_{\mathrm{BF16}}-L_{\mathrm{run}})/L_{\mathrm{BF16}}$, so negative is
  worse than BF16.  Triangles mark the off-axis native ablation value at
  step 7,500; Table~\ref{tab:quantized-validation-final} lists each path's
  activation-scale policy.}
  \label{fig:quantized-validation}
\end{figure}

\begin{table}[H]
  \centering
  \small
  \caption{Step-30,000 results on the fixed validation stream.  Lower NLL is
  better.  FP4 indicates whether the configured eligible linears execute with
  quantized weights and activations during inference.  Bold marks the lowest
  FP4 NLL; BF16 is the unquantized control.}
  \label{tab:quantized-validation-final}
  \begin{tabularx}{\textwidth}{@{}Ylcr@{}}
    \toprule
    Path & Activation scaling & FP4 & NLL \\
    \midrule
    BF16-trained trajectory, unquantized inference & not applicable & no & 2.27834 \\
    \uefp{} with Transformer Engine settings & current, $D=1$ & yes & \textbf{2.30376} \\
    Proposed \uefp{}, $B=16$, probe-matched & delayed, $D=50$ & yes & 2.32230 \\
    Transformer Engine \nv{} recipe & current, $D=1$ & yes & 2.32592 \\
    Proposed \uefp{}, $B=16$, decoded-operand Torch & delayed, $D=50$ & yes & 2.32900 \\
    Proposed \uefp{}, $B=32$, probe-matched & delayed, $D=50$ & yes & 2.33721 \\
    Native \nv{} no-RHT/all-linears ablation & current, $D=1$ & yes & 2.45468 \\
    \bottomrule
  \end{tabularx}
\end{table}

Among the FP4 paths, the proposed block-16 \uefp{} path has lower NLL than the
native Transformer Engine \nv{} recipe at 8 of 12 checkpoints, including all
four checkpoints from step 22,500 onward.  At step 30,000 it reaches 2.32230
versus 2.32592, a
difference of $-0.00362$ NLL.  Its NLL decreases from 2.42755 at step
22,500 to 2.32230 at step 30,000.  \uefp{} with Transformer Engine settings has
lower NLL than native Transformer Engine \nv{} at all 12 checkpoints and
finishes at 2.30376, a difference of
$-0.02217$ NLL.

Appendix~\ref{app:inference-policies} compares the three inference-scale
choices for the proposed block-16 checkpoint.  Their step-30,000 held-out NLLs
differ by less than $9.0\times10^{-5}$, and all remain below the native
Transformer Engine \nv{} result.

\subsection{Quantized downstream evaluation at step 30,000}
\label{sec:quantized-olmes}

We compare the seven final checkpoints on the OLMES likelihood-based
multiple-choice suite~\citep{gu2024olmes}.  We report its Core 9 aggregate,
MMLU~\citep{hendrycks2021mmlu}, and MMLU-Pro multiple-choice
accuracy~\citep{wang2024mmlupro}.

\begin{figure}[H]
  \centering
  \includegraphics[width=\textwidth]{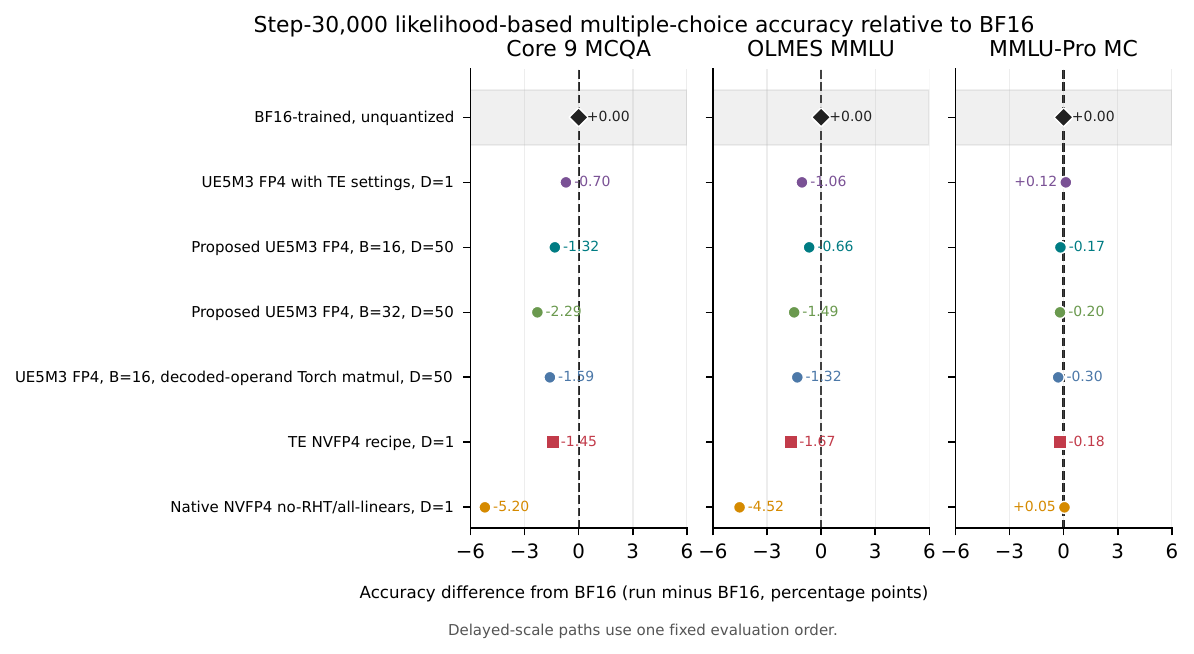}
  \caption{Step-30,000 likelihood-based multiple-choice accuracy relative to
  the separately trained BF16 control.  Values are percentage-point
  differences (run minus BF16), so positive values indicate higher accuracy.
  All quantized rows execute with their configured FP4 path active during
  inference.  The $D=50$ scores are fixed-order point estimates because delayed
  scale state depends on request order.}
  \label{fig:quantized-olmes}
\end{figure}

\begin{table}[H]
  \centering
  \small
  \caption{Step-30,000 OLMES accuracy.  Values are percentages; higher is
  better.  Bold marks the highest FP4 result in each column.  BF16 is the
  unquantized control.}
  \label{tab:quantized-olmes}
  \begin{tabularx}{\textwidth}{@{}Yllrrr@{}}
    \toprule
    Path & Inference precision & Scale policy & Core 9 & MMLU & MMLU-Pro MC \\
    \midrule
    BF16-trained trajectory & unquantized & not applicable & 65.14 & 39.64 & 11.24 \\
    \uefp{} with Transformer Engine settings & fake FP4 & current, $D=1$ & \textbf{64.43} & 38.57 & \textbf{11.36} \\
    Proposed \uefp{}, $B=16$, probe-matched & fake FP4 & delayed, $D=50$ & 63.82 & \textbf{38.97} & 11.06 \\
    Proposed \uefp{}, $B=32$, probe-matched & fake FP4 & delayed, $D=50$ & 62.85 & 38.14 & 11.04 \\
    Proposed \uefp{}, $B=16$, decoded-operand Torch & fake FP4 & delayed, $D=50$ & 63.54 & 38.32 & 10.94 \\
    Transformer Engine \nv{} recipe & native FP4 & current, $D=1$ & 63.69 & 37.96 & 11.05 \\
    Native \nv{} no-RHT/all-linears ablation & native FP4 & current, $D=1$ & 59.94 & 35.11 & 11.29 \\
    \bottomrule
  \end{tabularx}
\end{table}

Among the FP4 paths, relative to the native Transformer Engine \nv{}
checkpoint, proposed block-16 \uefp{} is higher by 0.13, 1.01, and 0.01
percentage points on Core 9, OLMES MMLU, and MMLU-Pro MC, respectively.
\uefp{} with Transformer Engine settings is higher by 0.74, 0.61, and 0.31
percentage points.

\subsection{Native \nv{} execution ablation}

The Blackwell/Transformer Engine FP4 path evaluated here exposes native \nv{}
with E4M3 block scales, not \ue{} block scales~\citep{nvidia2026tedocs}.  We
therefore measure the joint execution effect of removing RHT and moving 16
eligible final-block projections from BF16 to FP4.  Both native \nv{}
configurations use Transformer Engine current-tensor scaling ($D=1$), SR, and
two-dimensional weight scaling.

We measure synchronized forward and backward execution of the 8B model body on
one NVIDIA GB200 after warm-up.  Data loading, optimization, distributed
communication, and compilation are excluded from the timed region.

\begin{table}[H]
  \centering
  \small
  \caption{Native-\nv{} execution ablation.  ``100-step wall'' is the sum of
  synchronized model-body forward/backward timings after warm-up.}
  \label{tab:runtime}
  \begin{tabularx}{\textwidth}{@{}Yrrrr@{}}
    \toprule
    Recipe executed with native \nv{} & FP4 linears & BF16 linears & 100-step wall & Tokens/s \\
    \midrule
    Transformer Engine: RHT; 16 final-block linears in BF16
      & 96 & 16 & 31.877 s & 3,212 \\
    No RHT; all 112 eligible linears in FP4
      & 112 & 0 & \textbf{26.298 s} & \textbf{3,894} \\
    \midrule
    Relative change
      & +16 & $-16$ & -- & \textbf{$+21.2\%$} \\
    \bottomrule
  \end{tabularx}
\end{table}

The joint change raises measured model-body throughput from 3,212 to 3,894
tokens per second, an increase of 21.2\%.

\section{Conclusion}

This work demonstrates end-to-end software-emulated \uefp{} pretraining on a
Nemotron-H 8B model over 188.7 billion tokens.  Our proposed recipe combines
periodic sample-and-hold tensor scaling, two-dimensional weight scaling, and
upstream-gradient SR while omitting RHT and applying FP4 to all 112 eligible
internal linears instead of retaining a BF16 exemption for final-block
projections.  It achieves a lower final-window mean training loss than
NVIDIA's Transformer Engine \nv{} recipe.  \uefp{} with Transformer Engine
settings produces the lowest final-window training loss among the completed
FP4 trajectories.

Under their configured quantized-inference paths, the proposed block-16
\uefp{} path has lower held-out NLL on
the fixed validation stream than native Transformer Engine \nv{} at 8 of 12
checkpoints, including all four final checkpoints, and finishes at 2.32230
versus 2.32592.  \uefp{} with Transformer Engine settings has lower NLL at all
12 checkpoints and finishes at 2.30376.

Among the FP4 paths at step 30,000, the proposed block-16 \uefp{} point
estimates are higher than native Transformer Engine \nv{} by 0.13, 1.01, and
0.01 percentage points on Core 9, OLMES MMLU, and MMLU-Pro MC, respectively.
The \uefp{} path with Transformer Engine settings is higher by 0.74, 0.61, and
0.31 percentage points.

In the native \nv{} execution ablation, the no-RHT/all-linears configuration
records 21.2\% higher measured model-body token throughput than the Transformer
Engine configuration.
The probe-matched emulator also reproduces native model behavior in
deterministic controls, supporting its use for the \uefp{} experiments.

Because the evaluated Blackwell/Transformer Engine path supports E4M3 rather
than \ue{} block scales~\citep{nvidia2026tedocs}, these results motivate future
accelerators to support \ue{} block scaling directly and realize the \uefp{}
recipe in hardware.

\section*{Generative-AI Disclosure}

OpenAI Codex was used extensively to assist with code development, experiment
orchestration, analysis, programmatic figure generation, and drafting and
editing this report.  Each human author takes full responsibility for the code,
experimental design and execution, analysis, numerical results, figures,
citations, and text.

\bibliographystyle{plainnat}
\bibliography{references}

\clearpage
\appendix

\section{Inference-Scale Sensitivity}
\label{app:inference-policies}

For the proposed block-16 checkpoint, we compare the three policies summarized
in Table~\ref{tab:inference-scale-summary}.  In every case, each loaded weight
tensor's amax is measured once and its tensor-wide reference remains fixed.
The delayed cache starts from the first inference batch rather than restoring
training state.  The calibrated-frozen policy uses the largest activation amax
observed for each eligible linear over 64 calibration sequences that are
disjoint from validation.

Delayed inference is order-dependent, so validation and downstream evaluation
use fixed request orders.  For the downstream suite, the scale state advances
continuously rather than restarting at task boundaries.

\begin{table}[H]
  \centering
  \small
  \caption{Held-out NLL under three activation-scale
  policies for the proposed block-16 checkpoint.  Every populated cell uses
  fake-quantized FP4 weights and activations on the same validation stream.
  Lower is better; an em dash denotes a policy not
  evaluated at that checkpoint.}
  \label{tab:activation-scale-policies}
  \begin{tabular}{@{}rrrr@{}}
    \toprule
    Checkpoint step & Current tensor, $D=1$ & Delayed, $D=50$ & Calibrated frozen \\
    \midrule
    27,500 & --- & 2.364080 & 2.364100 \\
    30,000 & 2.322215 & 2.322305 & 2.322243 \\
    \bottomrule
  \end{tabular}
\end{table}

At step 30,000, the largest difference among the three policies is only
$8.99\times10^{-5}$ NLL, and each remains below the native Transformer Engine
\nv{} result of 2.325921.  The choice among them therefore does not change the
point-estimate ordering at this checkpoint.

\section{FP4 GEMM Simulation Details and Ablations}
\label{app:gemm-ablations}

We infer the hardware addition behavior using several deliberately different
probes, summarized in
Table~\ref{tab:gemm-probes}.  Here an \emph{edge case} means an output close to
a rounding boundary, not a NaN, infinity, overflow, or zero-sized input.

The seven primary witnesses are real scalar outputs from full
$8192\mathbin{\times}2048\mathbin{\times}6144$ \texttt{feed\_forward.w2}
GEMMs in layers 2, 5, 7, 8, 9, and 11.  They include five positive and two
negative outputs with magnitudes from $1.5\times10^{-5}$ to $1.16\times10^{-1}$.
For every witness, the custom and TE paths had identical decoded FP4 operand
values, used scale bytes, and global scale product.  The decoded-operand Torch
matmul nevertheless chose the adjacent BF16 value because
its unrounded result lay only $2.3\times10^{-9}$ to $2.2\times10^{-8}$ from the
midpoint between the two BF16 choices.  These deliberately separating cells
form a diagnostic witness set, on which the decoded-operand matmul matches 0/7 outputs.
Table~\ref{tab:gemm-witnesses} lists the complete set; coordinates are
zero-based and ``offset'' is the decoded-operand Torch result minus the BF16
midpoint.

\begin{table}[H]
  \centering
  \scriptsize
  \caption{The seven real-model BF16-boundary witnesses.  A positive midpoint
  offset lies above the boundary and a negative offset lies below it.}
  \label{tab:gemm-witnesses}
  \begin{tabular}{@{}lrrccc@{}}
    \toprule
    Module & Row & Column & Native BF16 & Decoded-operand Torch BF16 & Midpoint offset \\
    \midrule
    \texttt{L2.w2}  & 4,143 & 11    & $1.4591217\!\times\!10^{-4}$  & $1.4495850\!\times\!10^{-4}$  & $-1.6793\!\times\!10^{-8}$ \\
    \texttt{L5.w2}  & 7,919 & 240   & $1.5258789\!\times\!10^{-5}$  & $1.5377998\!\times\!10^{-5}$  & $+3.1214\!\times\!10^{-9}$ \\
    \texttt{L7.w2}  & 6,189 & 809   & $2.2094727\!\times\!10^{-2}$  & $2.1972656\!\times\!10^{-2}$  & $-1.8626\!\times\!10^{-8}$ \\
    \texttt{L7.w2}  & 4,184 & 558   & $-2.4199486\!\times\!10^{-5}$ & $-2.4318695\!\times\!10^{-5}$ & $-9.7316\!\times\!10^{-9}$ \\
    \texttt{L8.w2}  & 3,263 & 1,005 & $7.3909760\!\times\!10^{-5}$  & $7.4386597\!\times\!10^{-5}$  & $+1.3788\!\times\!10^{-8}$ \\
    \texttt{L9.w2}  & 1,141 & 252   & $-1.1621094\!\times\!10^{-1}$ & $-1.1572266\!\times\!10^{-1}$ & $+2.2352\!\times\!10^{-8}$ \\
    \texttt{L11.w2} & 8,181 & 1,658 & $7.6599121\!\times\!10^{-3}$  & $7.6293945\!\times\!10^{-3}$  & $-2.3283\!\times\!10^{-9}$ \\
    \bottomrule
  \end{tabular}
\end{table}

\begin{table}[H]
  \centering
  \scriptsize
  \caption{Probe classes used to fit and test the output model.  Each class
  isolates a distinct source of agreement.}
  \label{tab:gemm-probes}
  \begin{tabularx}{\textwidth}{@{}p{0.17\textwidth}p{0.32\textwidth}YY@{}}
    \toprule
    Probe & Construction & What it isolates & Pass condition \\
    \midrule
    Seven W2 witnesses
      & Real model cells whose decoded-operand Torch result is adjacent to native
        BF16; operands and scales already agree
      & Accumulation after quantization
      & Native final BF16 value and encoded product-lattice bin \\
    258 order tests
      & Reorder the 384 K-blocks of the discriminating 6,144-term witness while
        preserving the same products and exact dot product
      & Product grouping, traversal order, and cross-group rounding
      & Native FP32 grid bin for every permutation \\
    Same-sign thresholds
      & Construct positive-only and negative-only partial sums immediately
        around an FP32 rounding boundary
      & Nearest versus toward-zero addition without cancellation ambiguity
      & Correct side of every separating threshold \\
    Broad 512-cell probes
      & Sample full $128\mathbin{\times}256\mathbin{\times}6144$ packed-random
        and TE-quantized GEMMs
      & Behavior away from hand-selected BF16 boundaries
      & Exact native FP32 value and final BF16 value per sampled cell \\
    1.29B model probe
      & One complete 1.2B-model forward and backward pass
      & Composition across forward, data-gradient, and weight-gradient GEMMs
      & Logits, loss, and complete parameter-gradient vector \\
    \bottomrule
  \end{tabularx}
\end{table}

The discriminating witness is the negative output at layer 7, row 4,184,
column 558.
Native TE returns $-2.4199486\times10^{-5}$.  An exact decoded sum followed by
product-lattice canonicalization lands in bin $-3371$, and 64-product groups combined
with nearest rounding land in bin $-3372$.  The native result is bin $-3370$;
only toward-zero combination reaches that bin.  Permuting this same witness's
K-blocks produces the 258 order tests: the mathematical dot product remains
fixed, while native rounded outputs change with physical order.  This is the
case that separates the 6/7 and 7/7 candidate output models.

Table~\ref{tab:gemm-ladder} then builds the match one rule at a time.

\begin{table}[H]
  \centering
  \small
  \caption{Output-model ladder against native Transformer Engine FP4 GEMM.
  Each row adds one probe-supported rule.}
  \label{tab:gemm-ladder}
  \begin{tabularx}{\textwidth}{@{}lYYcc@{}}
    \toprule
    Software output model & Added rule & What it tests & Order tests & BF16 witnesses \\
    \midrule
    Decoded-operand Torch matmul
      & Decode FP4 values to FP32, call a standard matrix multiply, then
        return BF16
      & Quantization only; native addition ignored
      & -- & 0/7 \\
    Exact sum + canonicalization
      & Exact decoded dot product, then RNE-map to its $1/1024$ product lattice
      & Whether final product-lattice mapping alone explains TE
      & -- & 6/7 \\
    64-product groups + nearest
      & One partial per 64 products; nearest rounding between groups
      & Physical grouping alone
      & 149/258 & 6/7 \\
    64-product groups + toward zero
      & Same groups; round inexact cross-group additions toward zero
      & Complete tested output rule
      & \textbf{258/258} & \textbf{7/7} \\
    \bottomrule
  \end{tabularx}
\end{table}

\begin{figure}[H]
  \centering
  \includegraphics[width=0.94\textwidth]{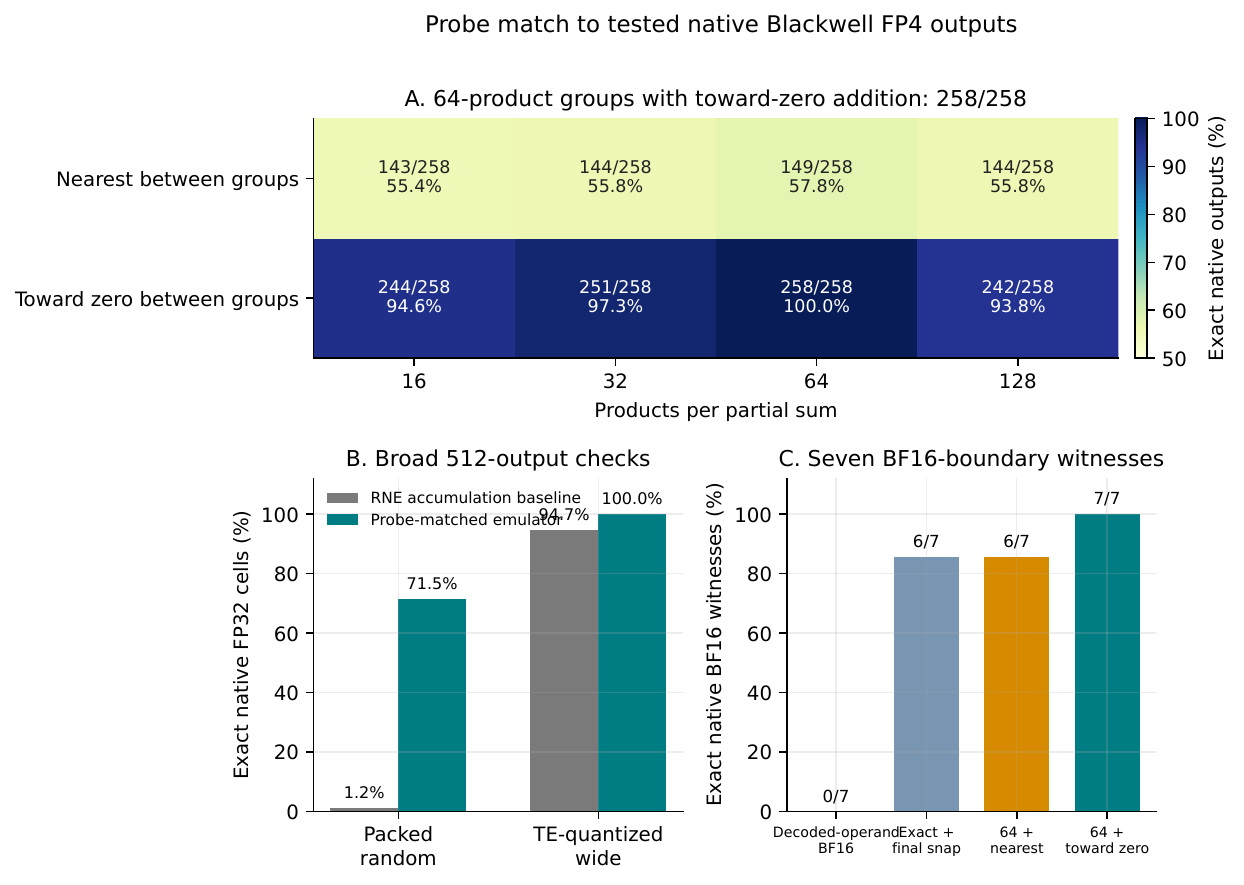}
  \caption{Probe evidence for the tested native Blackwell FP4 outputs.  (A)
  Reordering fixed products supports 64-product groups and toward-zero additions.
  (B) Broader samples compare exact native FP32 outputs.  (C) Seven sensitive
  BF16-boundary cases show which rules are necessary.  The final 1.2B check
  reproduces exact logits and loss and the complete 1,291,929,600-entry
  parameter-gradient vector in represented numerical value.}
  \label{fig:gemm}
\end{figure}

Figure~\ref{fig:gemm} adds two broader checks.  In panel A, rounding toward zero
matches more cases at every tested group size; groups of 64 are the only
258/258 match.  On 512 packed-random outputs in panel B, the complete model
matches 366 native FP32 values versus 6 for the RNE accumulation baseline.  On wide
TE-quantized operands it matches all 512, versus 485.  Both methods happen to
agree after BF16 conversion in these broad samples.  This is why a BF16-only
output comparison can miss the difference and why the sensitive cases matter.
On the seven selected witnesses, all 7/7 modeled outputs match the native final
BF16 values and encoded product-lattice bins; 5/7 also match the native
pre-BF16 FP32 value.

A third probe with TE quantization, RHT, and 2D weights gives 512/512 matches
for both methods.  In the denominator sweep of
Table~\ref{tab:grid-denominator-native}, $1/1024$, finer grids, and no final
canonicalization all match the 258 order tests, while the coarser grids fail.
The 1.2B test then matches every represented parameter-gradient value across
one complete model forward and backward pass.

\section{Statistical Consequences of the Probe-Matched Emulator}
\label{app:gemm-statistics}

We characterize the statistical consequences of the output model in
Equations~\ref{eq:issue-partial}--\ref{eq:final-grid} by holding the products
fixed and changing only two choices: how 64-product partial sums are combined
and whether the final encoded accumulator is canonicalized on the $1/1024$
product lattice.  The public NVIDIA Parallel Thread Execution (PTX)
instruction-set contract leaves E2M1 matrix-multiply-accumulate (MMA) order and
rounding unspecified.

The main sweep uses 4,096 independent dot products of length 4,096.  Inputs
come from a unit Gaussian, a Gaussian with standard deviation 32, a
variance-normalized Laplace distribution, a variance-normalized
Student-$t(3)$ distribution, and a contaminated Gaussian with 1\% of values
drawn with 25 times the base standard deviation.  We test both uncorrelated
operands and operands with correlation 0.25.  Every case receives identical
BF16 products under four accumulator variants:

\begin{center}
\begin{tabular}{@{}ll@{}}
  \toprule
  Variant & Operation \\
  \midrule
  RNE & Round-to-nearest-even FP32 cross-group additions \\
  RNE + grid & RNE, then nearest-even mapping to multiples of $1/1024$ \\
  RTZ & Round-toward-zero FP32 cross-group additions \\
  RTZ + grid & RTZ, then nearest-even mapping to multiples of $1/1024$ \\
  \bottomrule
\end{tabular}
\end{center}

For an exact addition result $s$, one toward-zero rounding error has the form
\begin{equation}
  e_{\mathrm{RTZ}} = \operatorname{RTZ}_{32}(s)-s
  = -\operatorname{sign}(s)\delta,
  \qquad 0 \leq \delta < \operatorname{ulp}(s).
  \label{eq:rz-contraction}
\end{equation}
Here $\operatorname{ulp}(s)$ is the spacing between adjacent FP32 values at
$s$, or one unit in the last place.
Thus an inexact positive addition moves downward and an inexact negative
addition moves upward: both move toward zero.  A symmetric distribution can
still have almost zero \emph{unconditional signed} error because its positive
and negative conditional biases cancel.  Magnitude bias and signal gain are
therefore more informative than mean signed error.

Table~\ref{tab:rz-statistics} confirms this contraction.  RNE accumulation is
within approximately 0.004 parts per million (ppm) of zero magnitude/gain bias
in the same cases.  Increasing Gaussian variance does not remove the relative
effect, and heavier tails make it slightly larger rather than changing its
sign.

\begin{table}[H]
  \centering
  \small
  \caption{Relative effect of RTZ for BF16 $K=4{,}096$ dots.  Magnitude bias uses
  uncorrelated operands; gain error uses correlation 0.25.  Negative values
  mean contraction toward zero.}
  \label{tab:rz-statistics}
  \begin{tabular}{@{}lrr@{}}
    \toprule
    Distribution & Magnitude bias (ppm) & Gain error (ppm) \\
    \midrule
    Gaussian & $-1.169$ & $-1.283$ \\
    Gaussian, $\sigma=32$ & $-1.136$ & $-1.281$ \\
    Laplace & $-1.185$ & $-1.285$ \\
    Student-$t(3)$ & $-1.188$ & $-1.311$ \\
    1\% contaminated Gaussian & $-1.293$ & $-1.350$ \\
    \bottomrule
  \end{tabular}
\end{table}

A second probe uses linear regression with 2,048 examples and 2,048 features.
The RTZ gradient retains cosine similarity of at least
$0.999999999996$ with the exact-accumulation gradient, but its gain errors are
$-1.065$, $-1.213$, and $-1.219$ ppm for Gaussian, Student-$t(3)$, and
contaminated-Gaussian inputs, respectively.  The corresponding RNE errors are
$+0.011$, $+0.013$, and $-0.004$ ppm.  RTZ therefore preserves direction while
attenuating amplitude by roughly 1 ppm.  In the correlated ($\rho=0.25$)
E2M1-plus-E4M3 controls, input quantization contributes 1.24--1.83\% relative
root-mean-square error, whereas changing RNE to RTZ affects only the sixth or
seventh decimal place.

The product-lattice canonicalization behaves differently.  With
$d=1024$ and $\Delta=d^{-1}=2^{-10}$, Equation~\ref{eq:grid-denominator}
becomes
\begin{equation}
  Q_{\Delta}(z)=\Delta\operatorname{RNE}(z/\Delta),
  \qquad |Q_{\Delta}(z)-z|\leq \Delta/2.
  \label{eq:grid-denoiser}
\end{equation}
It maps every encoded accumulator in $[-1/2048,+1/2048]$ to zero.  After the
tensor decode factor $\alpha$, the corresponding real-value dead-zone interval
is $[-\alpha/2048,+\alpha/2048]$, with total width $\alpha/1024$.
For 4,096 exact-cancellation trials at 64-product partial-sum scale 1, RNE accumulation
leaves root-mean-square (RMS) path-rounding residue of $6.34\times10^{-7}$ and RTZ leaves
$5.40\times10^{-6}$; either gridded variant returns exact zero in every trial.
When a genuine residual has standard deviation $1/4096$, however, the grid
maps 95.4\% of outputs to zero, lowers gain to approximately 0.44, and raises
root-mean-square error (RMSE) to $2.26\times10^{-4}$.  At 64-product partial-sum scale 32, even the RTZ path error
can exceed half a grid cell.

We then repeat the near-cancellation probe while varying $d$.  Each row in
Table~\ref{tab:grid-denominator-statistics} uses the same 4,096 Gaussian
partial-sum sequences and the same residual samples, so only the final grid
changes.  The true residual has standard deviation $1/4096$, and the
64-product partial-sum scale is one.  The expected tradeoff is visible directly: coarser grids
remove cancellation residue aggressively but also erase the intended signal;
finer grids preserve gain and reduce RMSE.

\begin{table}[H]
  \centering
  \small
  \caption{Effect of final-grid granularity on a genuine near-zero residual.
  ``None'' retains the RTZ result without final canonicalization.}
  \label{tab:grid-denominator-statistics}
  \begin{tabular}{@{}lrrr@{}}
    \toprule
    Grid & Mapped to zero (\%) & Signal gain & RMSE \\
    \midrule
    None     & 0.02  & 1.0002 & $5.42\times10^{-6}$ \\
    $1/256$  & 100.00 & 0.0000 & $2.47\times10^{-4}$ \\
    $1/512$  & 99.98 & 0.0083 & $2.46\times10^{-4}$ \\
    $1/1024$ & 95.43 & 0.4332 & $2.27\times10^{-4}$ \\
    $1/2048$ & 67.68 & 0.9964 & $1.41\times10^{-4}$ \\
    $1/4096$ & 37.92 & 1.0017 & $7.08\times10^{-5}$ \\
    \bottomrule
  \end{tabular}
\end{table}

\begin{figure}[H]
  \centering
  \includegraphics[width=0.98\textwidth]{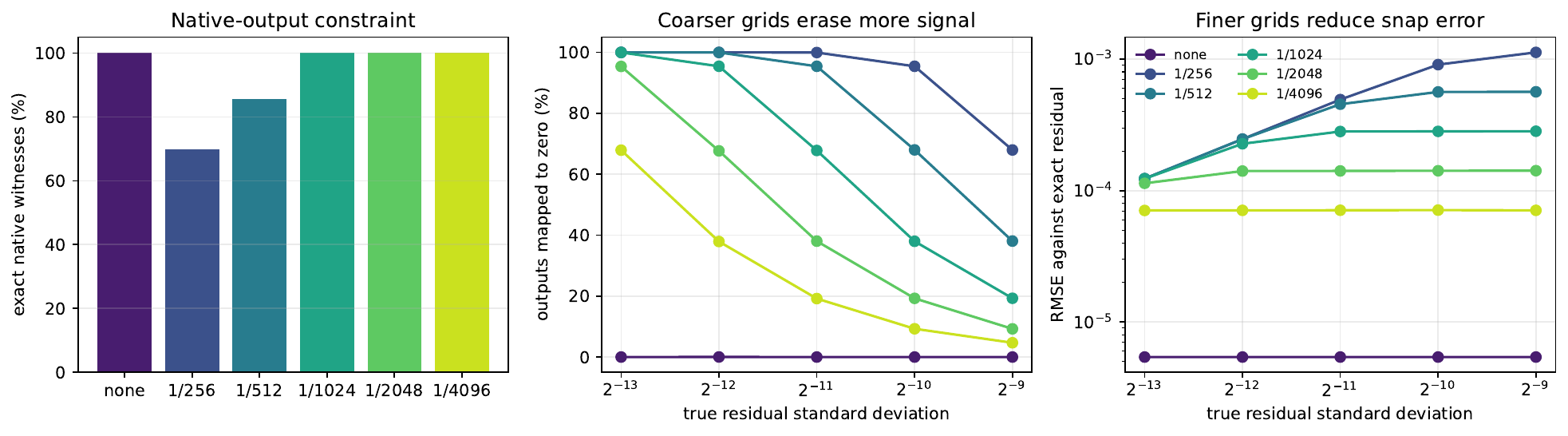}
  \caption{Final-grid granularity sweep.  Left: coarser $1/256$ and $1/512$
  grids fail native witnesses; $1/1024$, finer grids, and no snap all match the
  258 permutation witnesses.  Center and right: coarser grids map more real near-zero
  residuals to zero and incur larger error.}
  \label{fig:grid-denominator}
\end{figure}

\subsection{Optimizer-level regression control}

We first test whether the final-grid denominator changes optimization in a
compact two-layer teacher--student regression,
$256\rightarrow256\rightarrow64$, with
81,920 trainable parameters.  It uses 2,048 fixed training examples, 1,024
fixed evaluation examples, minibatches of 128, and 500 AdamW updates.  The
student uses the same E2M1/E5M3 quantization and probe-matched GEMM as the main
experiments for the forward, data-gradient, and weight-gradient matrix products.  We
set the scale target to 448, update delayed amax every 50 steps, disable RHT,
and change only $d$.

For each of seeds 41, 42, and 43, all grid choices receive the same teacher,
training set, initialization, minibatch order, and random stream.  We run the
experiment once with nearest-even gradient quantization and once with
stochastic gradient quantization.  In both cases, all 500 recorded losses,
gradient norms, and zero-gradient rates are exactly equal to the no-grid
trajectory.  After the last update, all 81,920 parameter values are also bit
exact.  Table~\ref{tab:grid-regression} counts the five nonzero denominators
against no grid over three seeds, hence 15 comparisons per row.

\begin{table}[H]
  \centering
  \small
  \caption{Tiny regression control.  Final-evaluation mean-squared-error (MSE)
  values are means over three seeds; exactness compares each nonzero grid with
  $d=0$.}
  \label{tab:grid-regression}
  \begin{tabular}{@{}lrrrr@{}}
    \toprule
    Gradient rounding & BF16 final-eval MSE & Quantized final-eval MSE & Exact records & Exact states \\
    \midrule
    Nearest-even & 0.001956 & 0.027130 & 15/15 & 15/15 \\
    Stochastic   & 0.001956 & 0.027578 & 15/15 & 15/15 \\
    \bottomrule
  \end{tabular}
\end{table}

\begin{figure}[H]
  \centering
  \includegraphics[width=0.98\textwidth]{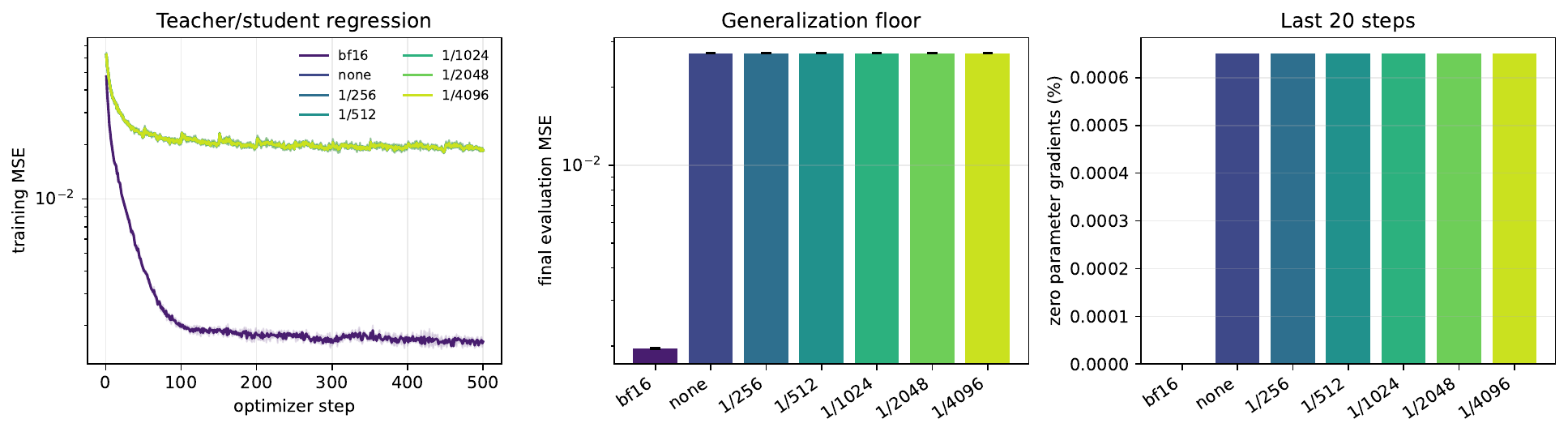}
  \captionsetup{width=0.92\textwidth}
  \caption{The 81,920-parameter regression control with nearest-even gradient
  quantization.  All six quantized curves and bars overlap exactly.  Repeating
  the experiment with stochastic gradient quantization gives the same grid
  invariance.}
  \label{fig:grid-regression}
\end{figure}

At the BF16 interface of this regression control, all tested final grids
produce identical values and optimizer trajectories through 500 updates.

\begin{figure}[H]
  \centering
  \includegraphics[width=0.94\textwidth]{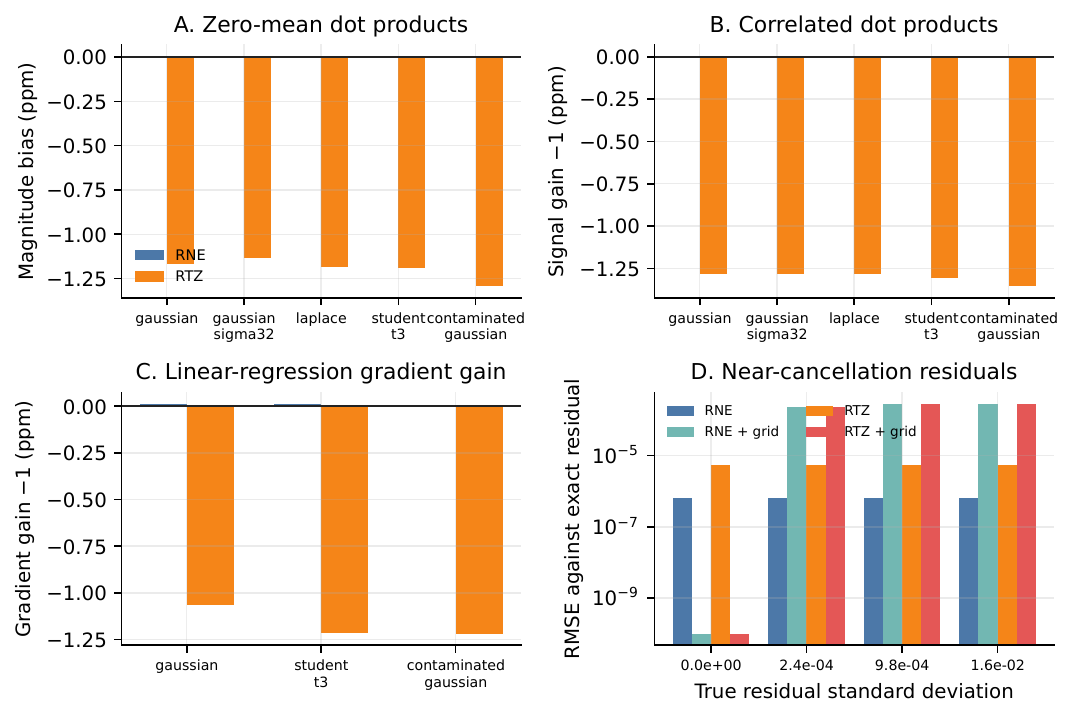}
  \caption{Statistical effect of the probe-matched emulator.  RTZ produces
  a small toward-zero gain contraction across Gaussian and heavy-tailed dot
  products and regression gradients.  Optional product-lattice
  canonicalization can eliminate exact-cancellation residue, but it also
  suppresses genuine residuals below its scale-dependent threshold.}
  \label{fig:issue-rz-statistics}
\end{figure}

\section{Hundred-Step Full-Model Native Parity}
\label{app:100step-parity}

The one-step model check in Figure~\ref{fig:gemm} establishes composed
forward/backward equality.  We extend the test through 100 AdamW updates to
compare evolving parameters and optimizer state.  The validator
uses the exact 1,291,929,600-parameter model: 20 layers, width 2,048, vocabulary
131,072, 96 quantized linears, and the final four transformer layers kept in
BF16.  Both trajectories use BF16 model tensors, batch size 1,
sequence length 64, RHT, 2D weight quantization, and no SR.  AdamW uses learning
rate $10^{-4}$, weight decay 0.1, and non-fused, non-foreach updates.

Native Transformer Engine and probe-matched trajectories run sequentially from one shared
initial state.  Step $s$ deterministically constructs a new synthetic token
batch from seed $5678+s-1$.  Before each update, the validator compares the
complete logits tensor and scalar loss.  After 100 updates, it scans every
gradient, parameter, and AdamW tensor state.  The native Transformer Engine
runtime is held fixed throughout the comparison.

\begin{table}[H]
  \centering
  \small
  \caption{Raw-storage parity after 100 full 1.292B-model AdamW updates.}
  \label{tab:100step-parity}
  \begin{tabularx}{\textwidth}{@{}lYr@{}}
    \toprule
    Object & Result & Compared values \\
    \midrule
    Per-step logits & Byte-exact at every step & 100/100 tensors \\
    Per-step losses & Byte-exact at every step & 100/100 scalars \\
    Final parameters & Byte-exact & 1,291,929,600 \\
    Final AdamW tensor states & Byte-exact & 2,583,859,363 \\
    Final gradients & One BF16 element differs by two units in the last place
      (ULPs) &
      1,291,929,600 \\
    \bottomrule
  \end{tabularx}
\end{table}

The complete gradient scan finds exactly one element that differs by two BF16
ULPs; the remaining 1,291,929,599 gradient elements are byte-exact.  This transient
difference does not alter the parameter update or either Adam moment: all
final parameters and optimizer tensor-state values remain byte-exact.

\section{Reproducibility Statement}

Code and reproduction instructions for the portable \uefp{} reference
implementation are available at
\url{https://github.com/MrHuff/ue5m3-fp4}.

\end{document}